\documentclass{article} 

\usepackage{iclr2027_conference,times}

\usepackage{amsmath,amsfonts,bm}

\def\eqref#1{equation~\ref{#1}}

\def\1{\bm{1}}

\DeclareMathAlphabet{\mathsfit}{\encodingdefault}{\sfdefault}{m}{sl}
\SetMathAlphabet{\mathsfit}{bold}{\encodingdefault}{\sfdefault}{bx}{n}

\usepackage{amsmath,amsfonts}
\usepackage{amsthm}
\usepackage{algorithmic}
\usepackage{algorithm2e}
\usepackage{graphicx}
\usepackage{textcomp}
\usepackage{xcolor}
\usepackage{multirow}
\usepackage{mathtools}
\usepackage{tcolorbox}
\usepackage{capt-of}
\usepackage{enumitem}
\usepackage{makecell}
\usepackage{float}
\usepackage{fancyhdr}
\usepackage{colortbl}
\usepackage{arydshln}
\usepackage{soul}
\usepackage{wrapfig}
\usepackage{titletoc}
\usepackage{pifont}
\usepackage{tikz}
\usepackage{booktabs}
\usepackage{listings}
\usepackage{hyperref}
\usepackage{url}

\RestyleAlgo{ruled}

\theoremstyle{plain}
\newtheorem{theorem}{Theorem}[section]

\title{
GrowPage: On-Demand KV Budgeting for Efficient LLM Reasoning Serving
}

\author{Qiankun Ma$^{1,2,3}$ \quad
Yanjiang Zhou$^{2}$ \quad
Zinan Xiong$^{2}$ \quad
Haofei Wang$^{2}$ \quad 
Zhen Song$^{2}$ \quad \\ 
\textbf{Yang Xiang}$^{2}$ \quad
\textbf{Ziyao Zhang}$^{2\dagger}$ \quad
\textbf{Hairong Zheng}$^{1,2,3\dagger}$
\\[2mm]
$^{1}$Shenzhen Institutes of Advanced Technology, Chinese Academy of Sciences
\\
$^{2}$Pengcheng Laboratory
\\
$^{3}$University of Chinese Academy of Sciences
\\[1mm]
\texttt{maqiankun25@mails.ucas.ac.cn}
\\[1mm]
{\small $^\dagger$Corresponding authors}
}

\iclrfinalcopy

\begin{document}

\maketitle

\pagestyle{plain}
\thispagestyle{plain}

\begin{abstract}
Long-output reasoning has made the key--value (KV) cache a critical memory bottleneck for efficient LLM serving.
Existing KV compression methods usually rely on a predefined per-request budget and adjust only which KV states are retained, leaving the total capacity fixed throughout decoding.
However, reasoning workloads exhibit substantial demand variation: different requests require different KV capacities, and the attention demand of an individual request evolves during generation.
We introduce \textbf{GrowPage}, an on-demand KV budgeting framework that treats KV capacity as a runtime resource.
GrowPage maintains lightweight dual-timescale query summaries to capture recent and long-term attention behaviors, and uses their relative attention working sets to estimate demand evolution.
At each capacity boundary, GrowPage either compresses KV states within the current allocation or acquires an additional physical page when broader demand emerges.
By integrating with PagedAttention's page-level memory abstraction, GrowPage preserves continuous batching and prefix caching.
Experiments on reasoning benchmarks across multiple models show that GrowPage achieves a superior performance--throughput trade-off over existing approaches.
\end{abstract}

\section{Introduction}
\label{sec:introduction}

Large reasoning models~\citep{yuan2025native,guo2025deepseek} achieve strong performance on challenging tasks such as mathematical reasoning and code generation by producing long chains of thought (CoTs) that iteratively explore, verify, and revise candidate solutions~\citep{wei2022chain}.
This capability, however, comes with substantial serving cost.
Reasoning models can generate thousands or even tens of thousands of tokens, continuously expanding their key--value (KV) caches.
When multiple long-running reasoning requests are served together, the resulting memory footprint limits the number of requests that can remain resident on GPU, making the KV cache a major bottleneck for efficient LLM serving.~\citep{hu2025raas,chitty2026pagedeviction}.

This pressure has motivated extensive research on KV cache compression.
Early methods mainly reduce prompt KV states~\citep{li2024snapkv,cai2024pyramidkv,feng2026ada}, while recent approaches extend token eviction to decoding to bound cache growth during long-output generation~\citep{ghadia2025dialogue,cai2026r,liao2025g,ramachandran2026thinkv}.
However, many such methods are difficult to integrate with modern serving mechanisms such as continuous batching and prefix caching, so their algorithmic memory savings do not necessarily translate into higher serving efficiency.
Zipage~\citep{liao2026zipage} bridges this gap through Compressed PagedAttention, combining token-wise KV eviction with PagedAttention while preserving key serving capabilities.

\begin{figure*}[t]
\centering
\includegraphics[width=0.90\linewidth,keepaspectratio]{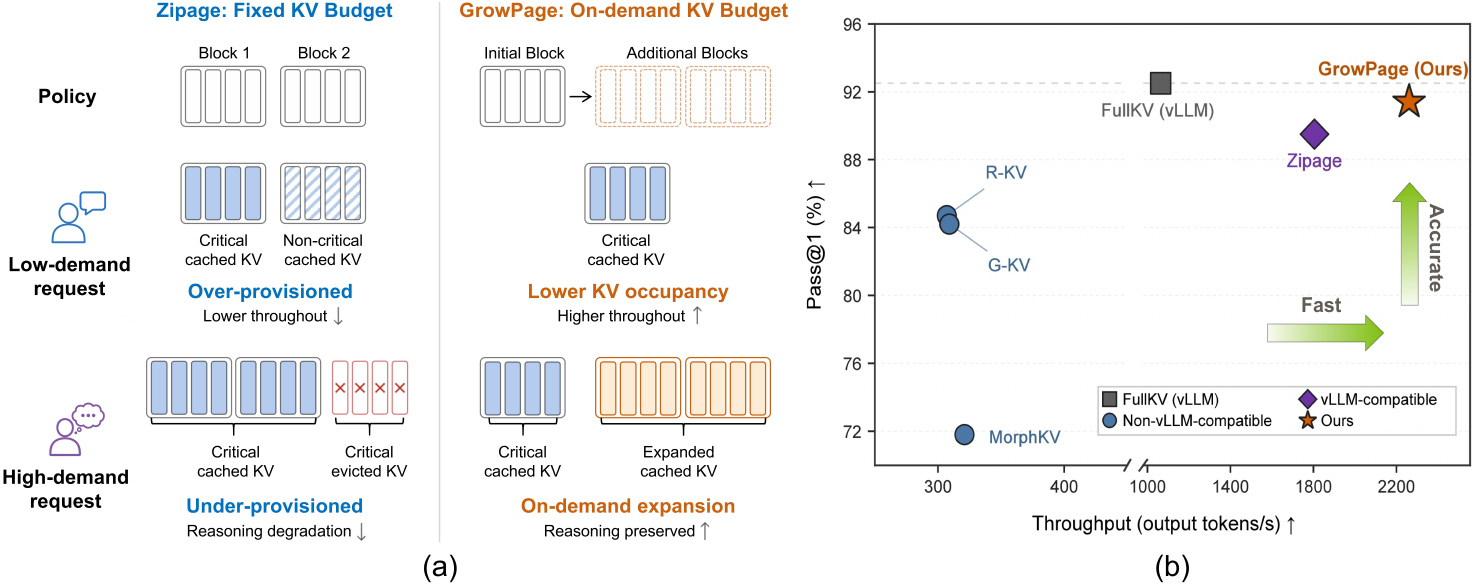}
\vspace{-4mm}
\caption{
Illustrative comparison of KV budgeting strategies for LLM reasoning.
(a) Fixed-budget Zipage and on-demand GrowPage for low- and high-demand requests.
(b) Reasoning performance vs. throughput on AMC23 with Qwen3-8B.
}
\vspace{-4mm}
\label{fig:overview}
\end{figure*}

Yet Zipage, like most existing KV compression methods, still operates under a predefined per-request KV capacity bound. 
This underutilizes PagedAttention's block-based memory management, which naturally supports incremental physical-page allocation, and mismatches the heterogeneous memory demands of reasoning requests.
Existing adaptive methods may change which KV states are retained or how a given budget is distributed, but typically leave the total per-request capacity fixed. GrowPage instead makes this capacity itself an online decision variable.
As illustrated in Figure~\ref{fig:overview}(a), a generous budget over-provisions low-demand requests, whereas an aggressive budget can under-provision demanding requests and evict reasoning-critical history.
Tuning a fixed budget therefore merely shifts the operating point between wasted memory and insufficient capacity.

Our analyses in Section~\ref{sec:dynamic_budget} reveal that this mismatch arises at two levels: the minimum sufficient KV budget varies substantially across requests, and attention concentration within an individual request also evolves throughout decoding.
These observations motivate treating per-request KV capacity as a runtime quantity rather than a static budget chosen in advance.
We theoretically connect attention concentration to the KV capacity required to preserve attention outputs, and empirically show that the relative working-set difference induced by recent and long-term query summaries provides a lightweight signal for tracking demand evolution.

Based on these insights, we introduce \textbf{GrowPage}, an on-demand KV budgeting framework for efficient LLM reasoning serving.
GrowPage maintains lightweight dual-timescale query summaries, using the long-timescale summary as a historical reference and the short-timescale summary to capture recent attention behavior.
Their relative working sets estimate the current demand trend.
At each capacity boundary, GrowPage either compresses historical KV states within the existing allocation or acquires an additional physical page.
By coupling this decision with PagedAttention memory management, GrowPage adapts per-request KV capacity while preserving continuous batching and prefix caching.
Across multiple mathematical reasoning and code-generation benchmarks and diverse model architectures, GrowPage consistently improves the reasoning performance--throughput trade-off; Figure~\ref{fig:overview}(b) shows a representative result on AMC23 with Qwen3-8B.

Our contributions are summarized as follows:
\begin{itemize}
[align=right,labelsep=2pt,labelwidth=1em,leftmargin=0.5em,nosep]

\item \textbf{We identify the demand mismatch underlying fixed KV budgeting in reasoning serving.}
KV requirements vary both across requests and throughout decoding, making static capacity bounds prone to either memory over-provisioning or insufficient retention.

\item \textbf{We propose GrowPage, an on-demand KV budgeting framework for reasoning LLMs.}
GrowPage uses dual-timescale query summaries to estimate evolving attention demand and guide online compression or incremental physical-page allocation.

\item \textbf{We co-design on-demand KV budgeting with PagedAttention-based serving.}
GrowPage translates online demand estimates into page-level memory actions while preserving continuous batching and prefix caching, yielding a better reasoning performance--throughput trade-off under LLM reasoning serving.

\end{itemize}

\section{Why On-Demand KV Budgeting?}
\label{sec:dynamic_budget}

Static KV budgeting assumes that memory demand can be adequately captured by a capacity fixed in advance for the entire decoding process. We examine this assumption from two complementary perspectives: inter-request heterogeneity and intra-request temporal variation.

\noindent\textbf{Observation 1: Minimum sufficient KV budgets vary substantially across requests.}
We first quantify the KV capacity required by individual reasoning requests.
For each request--budget pair, we perform multiple independent generations and aggregate their correctness to reduce sampling variance.
Given a set of budgets $\mathcal{B}$, we define the \emph{minimum sufficient KV budget} of request $i$ as:
\begin{equation}
B_i^{\star}
=
\min
\left\{
B\in\mathcal{B}
\;\middle|\;
c_i(B')=1,\ \forall B'\in\mathcal{B}, B'\ge B
\right\},
\label{eq:min_budget}
\end{equation}
where $c_i(B)$ indicates whether request $i$ is answered correctly under budget $B$, as determined from multiple independent generations.
Requiring correctness to hold under all larger tested budgets reduces the influence of occasional non-monotonic outcomes.
Requests that are not consistently solved under any tested budget are categorized as \emph{Unsolved}.

\begin{wrapfigure}{R}{0.40\textwidth}
\vspace{-6mm}
  \centering \includegraphics[width = 0.39\textwidth, keepaspectratio]{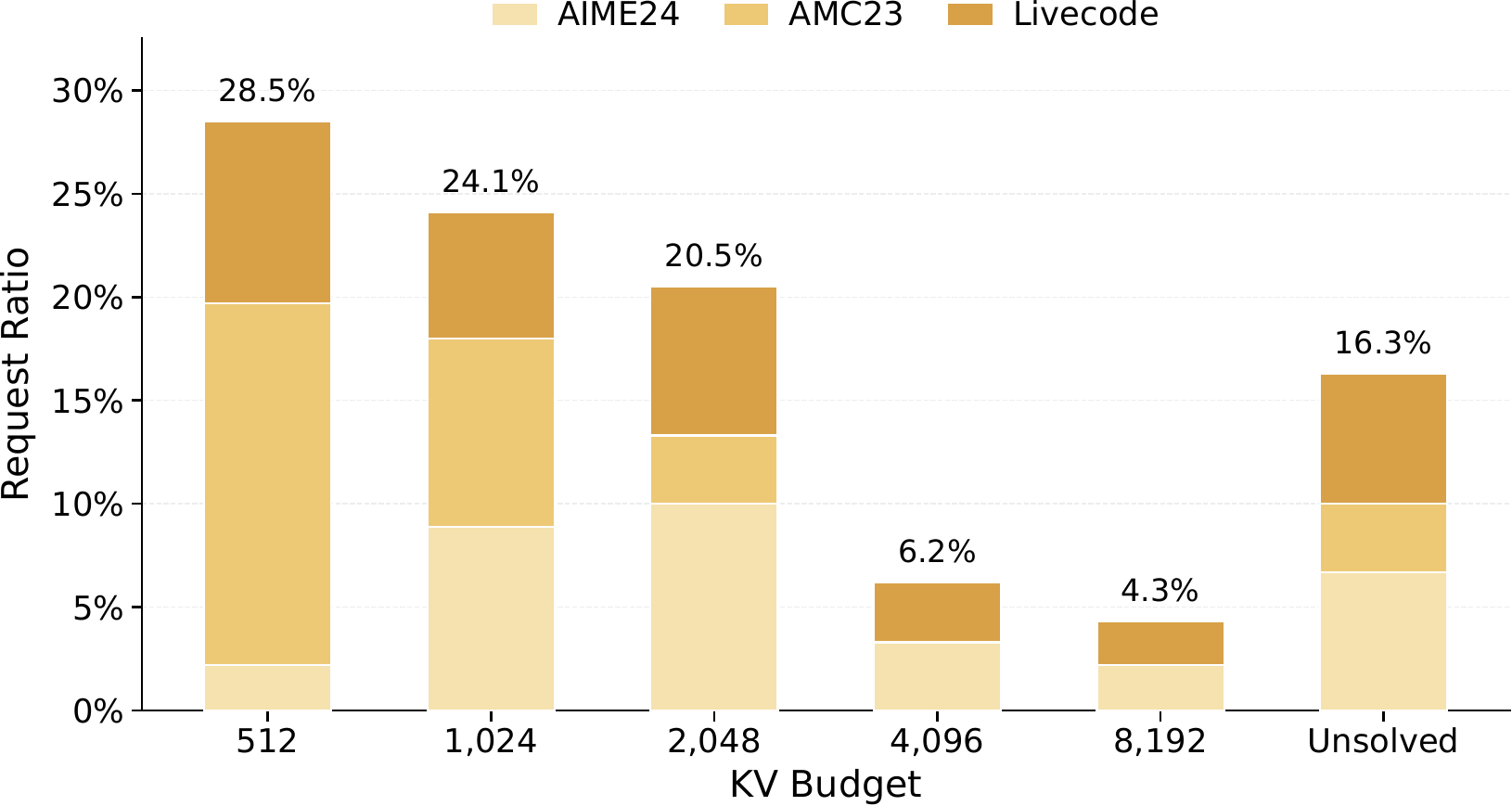}
  \vspace{-4mm}
  \caption{{Distribution of minimum sufficient KV budgets for Qwen3 8B.}}
  \vspace{-2mm}
  \label{fig:minimum_budgets}
\end{wrapfigure}

As shown in Figure~\ref{fig:minimum_budgets}, $B_i^\star$ spans the entire tested range across AIME24, AMC23, and LiveCode.
This wide distribution indicates substantial inter-request heterogeneity: a small fixed budget under-provisions demanding requests, whereas a large one unnecessarily reserves KV memory for requests requiring much less capacity.
Appendix~\ref{app:request_capacity_alignment} further shows that GrowPage generally allocates larger adaptive capacities to requests with higher minimum sufficient budgets, validating its request-level demand adaptation.
This inter-request heterogeneity already challenges the use of a single fixed budget. However, the mismatch is not limited to differences across requests. KV demand can also evolve over time within the same request.

\noindent\textbf{Observation 2: KV demand varies substantially during decoding.}
Prior work has shown that long-form reasoning traverses distinct reasoning stages with substantially different attention sparsity patterns~\citep{ramachandran2026thinkv}.

\begin{wrapfigure}{R}{0.40\textwidth}
\vspace{-6mm}
  \centering \includegraphics[width = 0.39\textwidth, keepaspectratio]{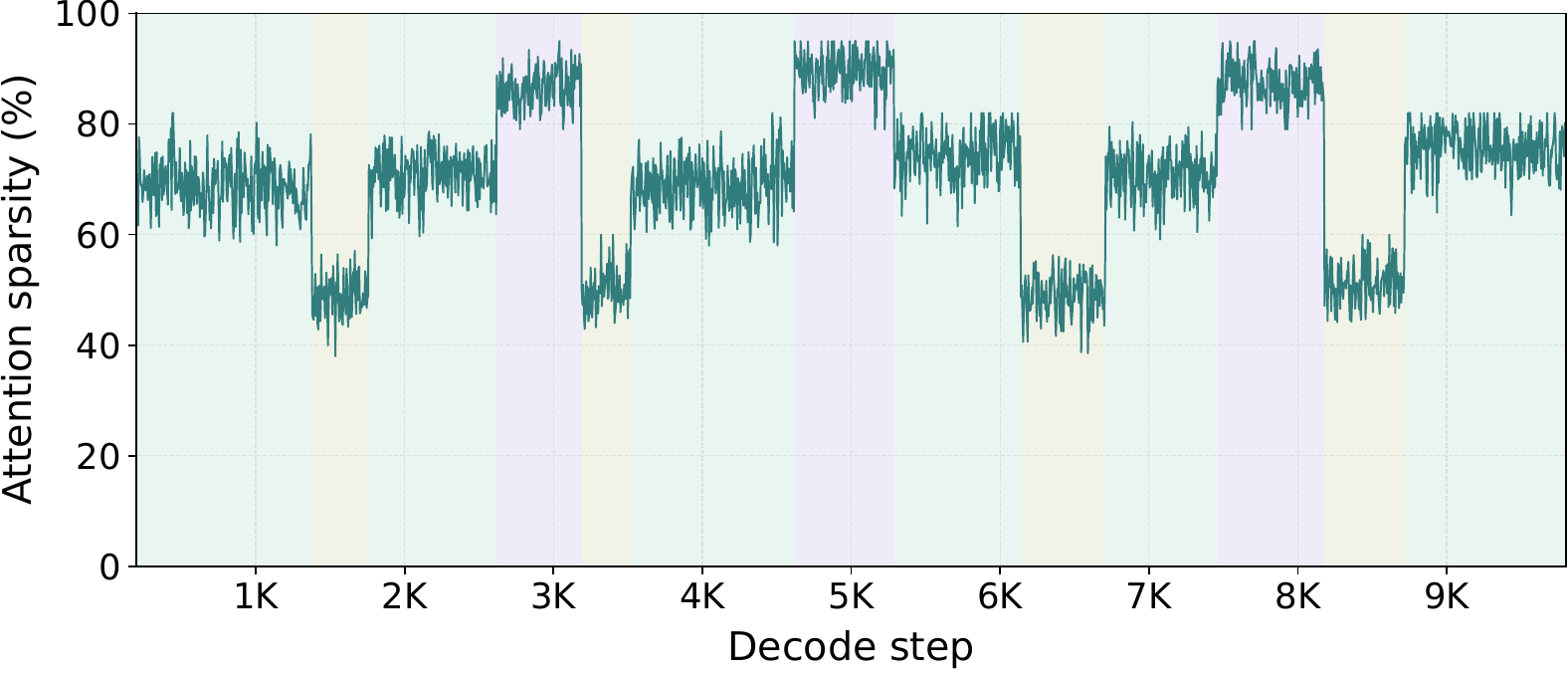}
  \vspace{-4mm}
  \caption{{Attention sparsity throughout decoding for DeepSeek-R1 Distill Llama 8B on LiveCodeBench.}}
  \vspace{-3mm}
  \label{fig:dynamic_demand}
\end{wrapfigure}

Consistent with this observation, Figure~\ref{fig:dynamic_demand} shows that the attention sparsity of a representative request changes throughout decoding, repeatedly transitioning between concentrated and diffuse regimes.
Such temporal variation suggests that the amount of historical KV information required by a request is itself non-stationary, making a capacity fixed before decoding unnecessarily conservative at some stages yet potentially insufficient at others.

Together, these observations reveal that KV demand varies both across requests and over the course of decoding.
They motivate treating KV capacity as an on-demand resource rather than a fixed reservation.
In the next section, we formalize the connection between attention concentration and the KV capacity required to preserve attention outputs, providing the theoretical basis for GrowPage's online capacity control.

\section{Methodology}
\label{sec:method}

\subsection{Theoretical Analysis}
\label{sec:theoretical_analysis}

Section~\ref{sec:dynamic_budget} shows that attention concentration varies substantially throughout decoding.
We now formalize why such variation implies changing KV capacity requirements.
At decoding step $t$, let
$\mathbf{a}_t=(a_{t,1},\ldots,a_{t,N_t})$
denote the normalized attention distribution over $N_t$ historical KV states, with
$a_{t,(1)}\ge\cdots\ge a_{t,(N_t)}$
denoting the sorted attention weights.
For a target coverage $p\in(0,1)$, we define the \emph{$p$-coverage attention demand} as:
\begin{equation}
D_p(\mathbf{a}_t)
=
\min
\left\{
k:
\sum_{i=1}^{k}a_{t,(i)}\ge p
\right\}.
\label{eq:attention_demand}
\end{equation}
Intuitively, concentrated attention yields a smaller $D_p$, whereas diffuse attention requires a larger working set.
We next connect this quantity to attention-output approximation.
Let $\mathbf{o}_t$ denote the full-cache attention output and
$\widehat{\mathbf{o}}_t(S)$ the output obtained by retaining and renormalizing a subset $S$:
\begin{equation}
\mathbf{o}_t
=
\sum_{i=1}^{N_t}a_{t,i}\mathbf{v}_i,
\qquad
\widehat{\mathbf{o}}_t(S)
=
\sum_{i\in S}
\frac{a_{t,i}}{m_t(S)}\mathbf{v}_i,
\quad
m_t(S)=\sum_{i\in S}a_{t,i}.
\label{eq:attention_outputs}
\end{equation}

Under the bounded-value assumption $\|\mathbf{v}_i\|_2\le V_{\max}$, we define the \emph{worst-case KV capacity}:
\begin{equation}
C_{\delta}^{\mathrm{wc}}(\mathbf{a}_t)
=
\min_{S}
\left\{
|S|:
\sup_{\|\mathbf{v}_i\|_2\le V_{\max}}
\left\|
\mathbf{o}_t-\widehat{\mathbf{o}}_t(S)
\right\|_2
\le \delta
\right\}.
\label{eq:worst_case_capacity}
\end{equation}
For any retained subset $S$, its worst-case approximation error is exactly:
\begin{equation}
\sup_{\|\mathbf{v}_i\|_2\le V_{\max}}
\left\|
\mathbf{o}_t-\widehat{\mathbf{o}}_t(S)
\right\|_2
=
2V_{\max}\bigl(1-m_t(S)\bigr).
\label{eq:worst_case_error}
\end{equation}
The upper bound follows from the triangle inequality, and is tight by assigning retained and discarded values to opposite directions with norm $V_{\max}$; the full proof is provided in Appendix~\ref{app:attention_capacity_proof}.

\begin{theorem}
\label{thm:attention_capacity}
Assume $\|\mathbf{v}_i\|_2\le V_{\max}$ for all $i$.
For any approximation tolerance $0<\delta<2V_{\max}$, the minimum worst-case KV capacity satisfies
$C_{\delta}^{\mathrm{wc}}(\mathbf{a}_t)=D_{1-\delta/(2V_{\max})}(\mathbf{a}_t)$.
\end{theorem}

Theorem~\ref{thm:attention_capacity} follows because guaranteeing error at most $\delta$ is equivalent to retaining attention mass
$m_t(S)\ge 1-\delta/(2V_{\max})$, and the maximum mass achievable with $k$ retained states is obtained by selecting the top-$k$ attention weights.
Thus, under the same tolerance, diffuse attention requires a larger robust KV capacity, whereas concentrated attention can be preserved with fewer states.
For the realized value vectors of a particular request, the required capacity may be smaller; Theorem~\ref{thm:attention_capacity} characterizes the worst-case requirement determined solely by the attention distribution.

Combined with the non-stationary attention behavior observed in Section~\ref{sec:dynamic_budget}, this result motivates adapting KV capacity as attention demand evolves during decoding.
We next describe how GrowPage tracks such changes online and translates them into on-demand capacity control.

\subsection{Online Demand Estimation}
\label{sec:demand_estimation}

\begin{figure*}[t]
  \centering
  \includegraphics[width=0.98\linewidth, keepaspectratio]{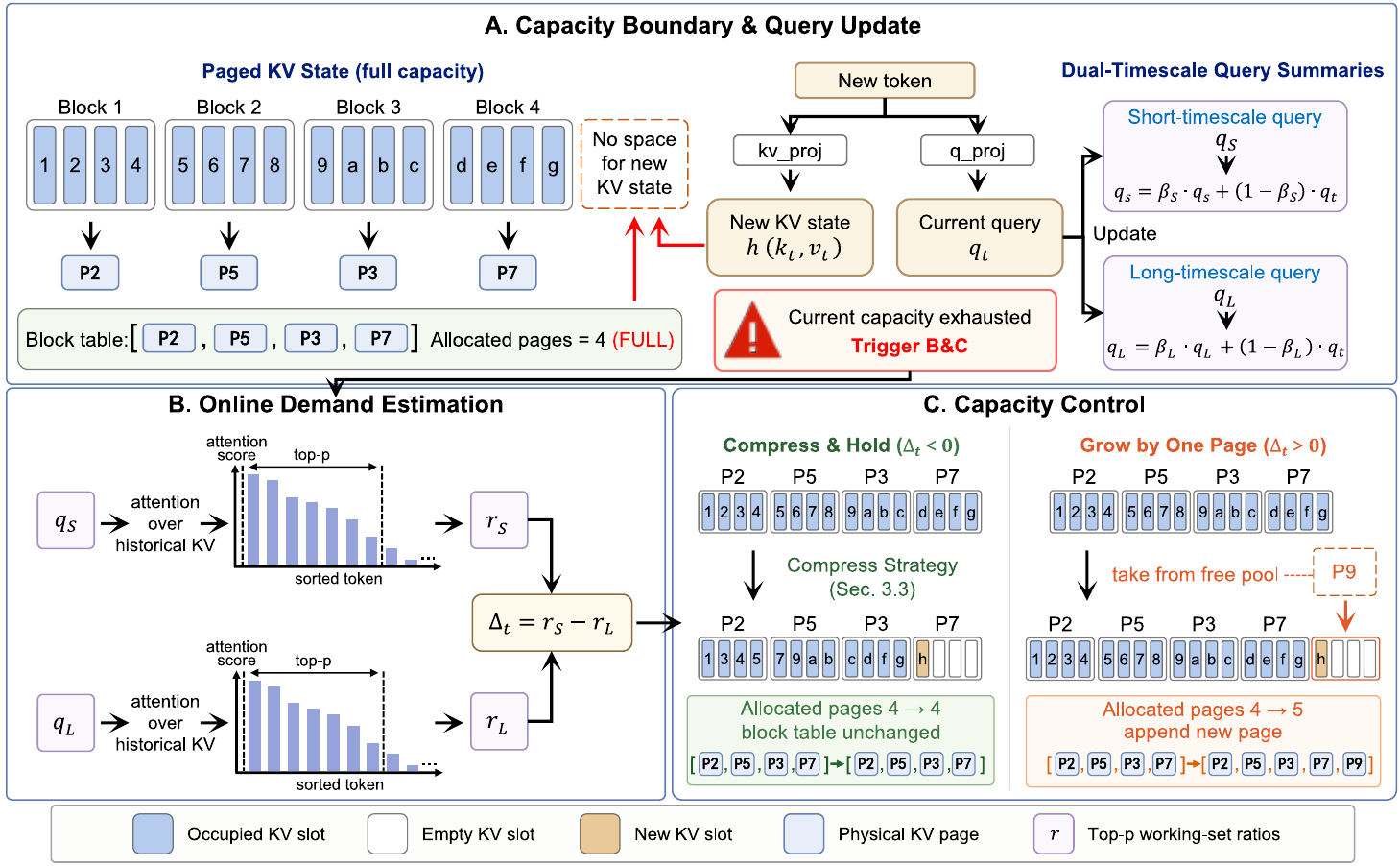}
  \vspace{-4mm}
  \caption{Overview of GrowPage. At each KV capacity boundary, GrowPage estimates the request's attention-demand trend and dynamically chooses between \emph{Compress \& Hold} and \emph{Grow by One Page}.}
  \vspace{-4mm}
  \label{fig:growpage_overview}
\end{figure*}

Figure~\ref{fig:growpage_overview} illustrates the overall workflow of GrowPage.
During decoding, each request maintains short- and long-timescale query summaries.
When the current KV capacity is exhausted, GrowPage compares their induced historical attention working sets to estimate the demand trend and decide whether to compress the existing KV states or acquire an additional physical page.

\paragraph{Dual-timescale query summaries.}
GrowPage maintains two exponentially smoothed query representations for each request and attention layer.
For clarity, we omit layer and head indices in Eq.~\ref{eq:dual_ema}.
Let $\mathbf{q}_t$ denote the current query after query normalization but before RoPE:
\begin{equation}
\bar{\mathbf{q}}_t^{x}
=
\beta_x\bar{\mathbf{q}}_{t-1}^{x}
+
(1-\beta_x)\mathbf{q}_t,
\qquad x\in\{S,L\},
\label{eq:dual_ema}
\end{equation}
where $\beta_S<\beta_L$; we use $\beta_S=0.9$ and $\beta_L=0.999$ by default.
The long-timescale summary serves as a historical reference, while the short-timescale summary captures recent attention behavior.
Their relative behavior therefore provides a lightweight estimate of demand evolution.
Both summaries are maintained in the pre-RoPE space to avoid averaging queries under different rotary transformations.
At each decision point, we calibrate their per-head RMS scales and apply RoPE at the current position before attention estimation.

\paragraph{Attention-demand trend.}
At a capacity boundary, GrowPage constructs a historical candidate set $\mathcal{C}_t$ and excludes the most recent KV block to reduce local-attention dominance.
For layer $\ell$ and KV head $h$, the aligned short- and long-timescale queries induce normalized attention distributions $\mathbf{a}_{t,\ell,h}^{S}$ and $\mathbf{a}_{t,\ell,h}^{L}$ over $\mathcal{C}_t$.
Following Section~\ref{sec:theoretical_analysis}, we compute
\begin{equation}
K_{t,\ell,h}^{x}
=
\min
\left\{
k:
\sum_{i=1}^{k}a_{t,\ell,h,(i)}^{x}\ge p
\right\},
\qquad
r_{t,\ell,h}^{x}
=
\frac{K_{t,\ell,h}^{x}}{|\mathcal{C}_t|},
\quad x\in\{S,L\},
\label{eq:working_set_ratio}
\end{equation}
where $a_{t,\ell,h,(i)}^{x}$ denotes the attention weights sorted in descending order, and we set $p=0.99$ by default.
Since page growth is a request-level decision, we follow the mean-aggregation convention of prior KV-cache methods and aggregate the working-set ratios across layers and KV heads:
\begin{equation}
r_t^{x}
=
\frac{1}{\sum_{\ell\in\mathcal{L}}|\mathcal{H}_{\ell}|}
\sum_{\ell\in\mathcal{L}}
\sum_{h\in\mathcal{H}_{\ell}}
r_{t,\ell,h}^{x},
\qquad
\Delta_t=r_t^{S}-r_t^{L},
\label{eq:demand_aggregation}
\end{equation}
where $\mathcal{L}$ denotes the participating layers and $\mathcal{H}_{\ell}$ the KV heads in layer $\ell$.
A negative $\Delta_t$ indicates increasing attention concentration, whereas a positive value indicates an expanding working set.
To translate this trend into a page-level decision, GrowPage compares $\Delta_t$ with a demand threshold $\tau$: it selects \emph{Grow by One Page} when $\Delta_t>\tau$, and otherwise performs \emph{Compress \& Hold}.
We use $\tau=0$ by default; its sensitivity is analyzed in Appendix~\ref{app:demand_threshold}.

\subsection{Trend-Guided Capacity Control}
\label{sec:capacity_control}

GrowPage invokes capacity control only when the current KV allocation is exhausted.
At each boundary, it chooses between \emph{Compress \& Hold} and \emph{Grow by One Page}, naturally aligning adaptation with PagedAttention's page granularity.
GrowPage adopts monotonic capacity growth rather than actively shrinking live requests, since page reclamation requires additional irreversible KV eviction; bidirectional resizing is studied in Appendix~\ref{app:capacity_policy_ablation}.

\paragraph{Compress \& Hold.}
When recent demand becomes more concentrated, GrowPage retains the current allocation and creates space by compacting historical KV states.
Unlike the request-level demand signal, KV selection is performed independently for each layer and KV head.
For candidate token $i$ at layer $\ell$ and head $h$, we combine its importance under the two temporal summaries as:
\begin{equation}
s_{i,\ell,h}
=
\max
\left(
a_{i,\ell,h}^{S},
a_{i,\ell,h}^{L}
\right).
\label{eq:token_score}
\end{equation}
Thus, different layer--head pairs can retain different token subsets, while the maximum across temporal views preserves tokens important to either recent or long-term attention.
GrowPage releases exactly one page KV states at each compression event.
Let $P_t$ denote the current number of allocated pages, $b$ the page size, and $\mathcal{R}_t$ the protected recent tokens.
The resulting historical retention budget is therefore determined by the target capacity $(P_t-1)b$, rather than treated as a tunable hyperparameter.
We denote this derived budget by $K_t$.
To prevent compression from over-concentrating on a few historical regions, GrowPage combines block-local retention with global selection independently for each layer--head pair.
Let $\mathcal{B}_j$ denote the candidate positions from historical block $j$ and $k_{\mathrm{loc}}$ the minimum local retention quota.
For each $(\ell,h)$,
\begin{equation}
\begin{aligned}
S_{\mathrm{loc}}^{\ell,h}
&=
\bigcup_j
\operatorname{TopK}_{k_{\mathrm{loc}}}
(\mathcal{B}_j; s_{\ell,h}),\\
S^{\ell,h}
&=
S_{\mathrm{loc}}^{\ell,h}
\cup
\operatorname{TopK}_{K_t-|S_{\mathrm{loc}}^{\ell,h}|}
\bigl(
\mathcal{C}_t\setminus S_{\mathrm{loc}}^{\ell,h};
s_{\ell,h}
\bigr),
\end{aligned}
\label{eq:local_global_selection}
\end{equation}
where $s_{\ell,h}=\{s_{i,\ell,h}\}_i$.
Thus, $k_{\mathrm{loc}}$ controls only the balance between block-local coverage and global importance, while the total number of retained historical tokens is fixed by the requirement to create one page worth of free slots.
The effect of $k_{\mathrm{loc}}$ is studied in Appendix~\ref{app:local_retention}.
The retained KV states are then compacted within the existing physical-page allocation.

\paragraph{Grow by One Page.}
When recent demand becomes more diffuse, GrowPage acquires one additional physical KV page instead of further compressing the historical working set.
The page is appended to the request's block table, and decoding continues until the next capacity boundary.

\section{System Integration}
\label{sec:system}

Section~\ref{sec:method} describes how GrowPage estimates the evolving KV demand of each request and converts it into page-level capacity decisions.
We now describe how these decisions are integrated into a PagedAttention-based serving runtime with low additional state overhead and minimal disruption to batched scheduling.

\subsection{Lightweight Per-Request State Management}
\label{sec:state_management}

GrowPage introduces only a small amount of auxiliary state for online demand estimation.
For each active request, the runtime maintains the short- and long-timescale query summaries described in Section~\ref{sec:demand_estimation}.
These states are allocated together with the request metadata, updated in place during decoding, and recycled when the request completes.
Unlike the KV cache itself, their size is independent of sequence length, avoiding additional memory growth as reasoning proceeds.

To avoid frequent GPU-side memory allocation during continuous batching, GrowPage pre-allocates the storage required by these summaries for active request slots and reuses it across requests.
We slightly modify the attention kernel to update the short- and long-timescale summaries from the normalized pre-RoPE queries during the regular decoding path.
The more expensive attention-based demand estimation is invoked only at capacity boundaries, so most decoding steps require only lightweight in-place state updates.

\subsection{Scheduler-Aware Page Allocation and Fallback}
\label{sec:scheduler_integration}

GrowPage maps the capacity decisions in Section~\ref{sec:capacity_control} directly onto PagedAttention's physical-page abstraction.
For \emph{Compress \& Hold}, the request keeps its existing physical-page allocation while selected historical KV states are compacted to release slots for subsequent tokens.
For \emph{Grow by One Page}, the scheduler obtains an additional page from the global free-page pool and appends its physical page identifier to the request's block table.
The request can then continue decoding without changing the underlying attention or block-table interface.

Page growth is coordinated with global memory availability.
When a free physical page is available, the growth request is satisfied immediately.
Under memory pressure, however, granting additional capacity to one request may reduce the memory available to other concurrent requests.
GrowPage therefore falls back to compression when additional page allocation cannot be safely satisfied, allowing the request to continue within its current allocation.
If global memory pressure remains unresolved, the existing serving scheduler can further apply controlled request preemption and rescheduling.
This keeps per-request adaptation subordinate to the global memory manager rather than allowing individual requests to monopolize the shared KV pool.

Importantly, GrowPage preserves the page/block abstraction of PagedAttention rather than introducing a separate memory layout.
It therefore remains compatible with serving mechanisms such as continuous batching, prefix caching, and CUDA Graph execution.
More details see Appendix~\ref{app:system_implementation}.
GrowPage additionally inherits the asynchronous compression pipeline of Zipage~\citep{liao2026zipage}, allowing KV compaction to overlap with serving execution and reducing compression-induced stalls on the critical decoding path.
Together, these design choices allow online KV-capacity adaptation to be incorporated into PagedAttention-based serving without sacrificing the system optimizations of the underlying engine.

\section{Experiments}
\label{sec:experiments}

\subsection{Experimental Setup}
\label{sec:setup}

\paragraph{Models and benchmarks.}
We evaluate GrowPage on two representative reasoning models:
DeepSeek-R1-Distill-Llama-8B~\citep{guo2025deepseek} and Qwen3-8B~\citep{yang2025qwen3}.
Following previous KV-cache compression studies, we consider both mathematical reasoning and code generation tasks, including GSM8K~\citep{cobbe2021training}, MATH500~\citep{hendrycks2021measuring}, AMC23, AIME24, and LiveCodeBench~\citep{jain2025livecodebench}.
These benchmarks cover diverse reasoning workloads with varying output lengths and KV-cache demands.

\begin{table*}[t]
\centering
\caption{
Main comparison on reasoning benchmarks.
We report pass@1 accuracy (\%) and decoding throughput (tokens/s). The results are evaluated on DeepSeek-R1-Distill-Llama-8B and Qwen3-8B.
}
\label{tab:main_results}

\setlength{\tabcolsep}{4pt}
\small

\resizebox{1.0\linewidth}{!}{
\begin{tabular}{lcccccccccccc}
\toprule
& \multicolumn{6}{c}{Pass@1 Accuracy (\%) $\uparrow$}
& \multicolumn{6}{c}{TPS (tokens/s) $\uparrow$}
\\
\cmidrule(lr){2-7}
\cmidrule(lr){8-13}

Method
& GSM8K
& MATH500
& AMC23
& AIME24
& LiveCode
& Avg.
& GSM8K
& MATH500
& AMC23
& AIME24
& LiveCode
& Avg.
\\
\midrule

\multicolumn{13}{l}{\textit{DeepSeek-R1-Distill-Llama-8B}}
\\
\midrule

FullKV (vLLM)
& 88.2 & 86.6 & 81.3 & 43.5 & 46.5 & 69.2
& 2705 & 1515 & 1235 & 766 & 1142 & 1472
\\

FullKV (nano-vLLM)
& 88.0 & 86.2 & 80.9 & 43.2 & 46.1 & 68.9
& 2576 & 1442 & 1182 & 722 & 1086 & 1401
\\

MorphKV \texttt{{\scriptsize (ICML'25)}}
& 85.5 & 75.8 & 70.4 & 35.8 & 33.9 & 60.3
& 507 & 235 & 189 & 119 & 178 & 245
\\

R-KV \texttt{{\scriptsize (NeurIPS'25)}}
& 87.1 & 84.7 & 77.8 & 39.9 & 42.2 & 66.3
& 450 & 218 & 177 & 102 & 162 & 221
\\

G-KV \texttt{{\scriptsize (arXiv'25)}}
& 87.7 & 85.1 & 78.3 & 40.5 & 42.8 & 66.9
& 472 & 228 & 185 & 111 & 171 & 233
\\

Zipage \texttt{{\scriptsize (ACL'26)}} 
& 88.0 & 85.0 & 79.0 & 41.4 & 44.6 & 67.6
& 3018 & 2174 & 2060 & 1158 & 1271 & 1936
\\

\cellcolor[HTML]{D5E8D4}\textbf{GrowPage (Ours)}
&\cellcolor[HTML]{D5E8D4}88.0 &\cellcolor[HTML]{D5E8D4}85.8 & \cellcolor[HTML]{D5E8D4}79.4 & \cellcolor[HTML]{D5E8D4}44.7 & \cellcolor[HTML]{D5E8D4}45.4 & \cellcolor[HTML]{D5E8D4}\textbf{68.7}
& \cellcolor[HTML]{D5E8D4}\textbf{3833} & \cellcolor[HTML]{D5E8D4}\textbf{2599} & \cellcolor[HTML]{D5E8D4}\textbf{2398} & \cellcolor[HTML]{D5E8D4}\textbf{1465} & \cellcolor[HTML]{D5E8D4}\textbf{1794} & \cellcolor[HTML]{D5E8D4}\textbf{2417}
\\

\midrule

\multicolumn{13}{l}{\textit{Qwen3-8B}}
\\
\midrule

FullKV (vLLM)
& 95.7 & 96.1 & 92.5 & 75.5 & 83.9 & 88.7
& 2658 & 1382 & 1064 & 607 & 1076 & 1357
\\

FullKV (nano-vLLM)
& 95.6 & 95.8 & 92.1 & 75.0 & 83.5 & 88.4
& 2486 & 1303 & 995 & 568 & 1015 & 1273
\\

MorphKV \texttt{{\scriptsize (ICML'25)}}
& 94.2 & 82.5 & 71.8 & 57.5 & 64.8 & 74.2
& 463 & 425 & 321 & 168 & 330 & 341
\\

R-KV \texttt{{\scriptsize (NeurIPS'25)}}
& 95.2 & 91.6 & 84.7 & 69.4 & 78.0 & 83.8
& 448 & 396 & 307 & 151 & 305 & 321
\\

G-KV \texttt{{\scriptsize (arXiv'25)}}
& 95.1 & 92.5 & 84.2 & 68.9 & 78.6 & 83.9
& 424 & 408 & 309 & 154 & 317 & 322
\\

Zipage \texttt{{\scriptsize (ACL'26)}}
& 95.7 & 93.7 & 89.5 & 72.7 & 80.7 & 86.5
& 3107 & 2076 & 1805 & 1007 & 1271 & 1853
\\

\cellcolor[HTML]{D5E8D4}\textbf{GrowPage (Ours)}
&\cellcolor[HTML]{D5E8D4}95.9 & \cellcolor[HTML]{D5E8D4}94.9 & \cellcolor[HTML]{D5E8D4}91.4 & \cellcolor[HTML]{D5E8D4}73.8 & \cellcolor[HTML]{D5E8D4}81.5 & \cellcolor[HTML]{D5E8D4}\textbf{87.5}
& \cellcolor[HTML]{D5E8D4}\textbf{3280} & \cellcolor[HTML]{D5E8D4}\textbf{2553} & \cellcolor[HTML]{D5E8D4}\textbf{2261} & \cellcolor[HTML]{D5E8D4}\textbf{1190} & \cellcolor[HTML]{D5E8D4}\textbf{1589} & \cellcolor[HTML]{D5E8D4}\textbf{2174}
\\

\bottomrule
\end{tabular}
}

\vspace{-3mm}
\end{table*}

\paragraph{Evaluation metrics and implementation.}
All experiments are conducted on a single NVIDIA A100 GPU with 80\,GB memory.
Following Zipage, we implement GrowPage on nano-vLLM with PagedAttention-based KV management and integrate the proposed capacity control mechanism into the decoding pipeline.
Unless otherwise specified, we follow the same system configuration as Zipage, fixing the KV block size $b$ to 256 tokens and the protected recent-window size $\mathcal{R}_t$ to 16 tokens.
We fix the GPU memory utilization ratio to 0.9 for all vLLM/nano-vLLM-based experiments to ensure a consistent memory budget.
We measure serving efficiency using decoding throughput (TPS), defined as the total number of output tokens generated across all requests divided by the wall-clock time from the start of decoding until all requests complete.
Additional experimental details, including implementation settings and dataset configurations, are provided in Appendix~\ref{app:dataset_details}.

\subsection{Main Results}
\label{sec:main_results}

\begin{wrapfigure}{R}{0.40\textwidth}
\vspace{-6mm}
  \centering \includegraphics[width = 0.39\textwidth, keepaspectratio]{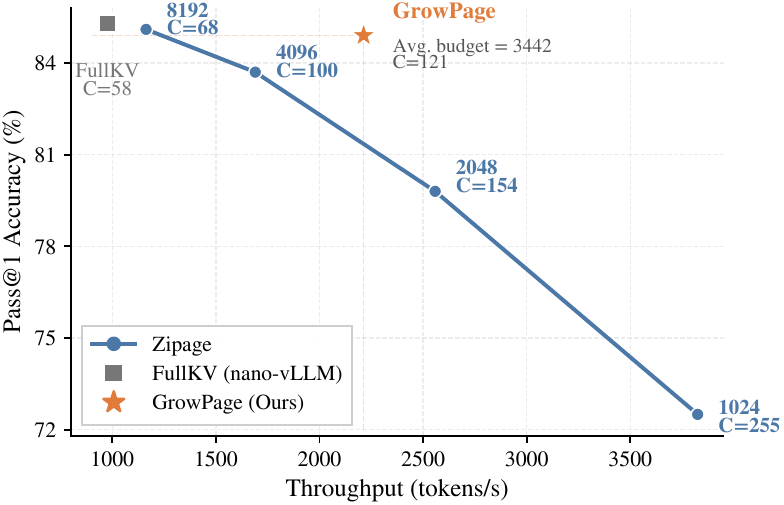}
  \vspace{-4mm}
  \caption{
Accuracy--throughput trade-off on the mixed GSM8K, AMC23, and AIME24 workload with Qwen3-8B.
}
  \vspace{-3mm}
  \label{fig:mixed_pareto}
\end{wrapfigure}
\paragraph{Benchmark comparison.}
Table~\ref{tab:main_results} compares GrowPage with full KV inference and representative KV-cache compression methods.
For fixed-budget baselines, we use the dataset-specific KV budgets reported in Table~\ref{tab:dataset_settings}.
To keep the main comparison concise, Table~\ref{tab:main_results} reports only Pass@1 accuracy and decoding throughput (TPS), while additional serving metrics and detailed analysis are provided in Appendix~\ref{app:main_results_analysis}.
All serving-system baselines use the same GPU memory budget, while MorphKV, R-KV, and G-KV use the largest executable batch size under the same memory constraint to maximize throughput.
GrowPage consistently achieves a better reasoning performance--throughput trade-off across both models.
Compared with full KV inference, GrowPage improves the average TPS by 64.3\% on DeepSeek-R1-Distill-Llama-8B and 60.2\% on Qwen3-8B while maintaining comparable pass@1 accuracy.
Although MorphKV, R-KV, and G-KV substantially reduce KV memory, their lack of serving-system co-design limits the resulting throughput gains even under maximized batching.
Compared with Zipage, which preserves PagedAttention compatibility, GrowPage further improves throughput by adapting KV capacity to request-specific demand rather than relying on a fixed budget.

\paragraph{Pareto Analysis under Mixed Workloads.}
To evaluate on-demand KV budgeting under heterogeneous request demands, we construct a mixed workload combining GSM8K, AMC23, and AIME24, and compare GrowPage with FullKV and Zipage under different KV budgets.
Details are provided in Appendix~\ref{app:mixed_workload}.
As shown in Figure~\ref{fig:mixed_pareto}, $C$ denotes the average number of concurrent requests resident in GPU memory during serving.
Increasing the fixed budget of Zipage improves accuracy but reduces $C$, resulting in lower throughput.
GrowPage achieves 84.9\% pass@1 accuracy at 2213 tokens/s with an average KV budget of 3442 and $C=121$.
It outperforms Zipage-4096 in accuracy (84.9\% vs.\ 83.7\%), throughput (2213 vs.\ 1690 tokens/s), and resident concurrency ($121$ vs.\ $100$), while approaching the accuracy of Zipage-8192 (85.1\%) with nearly $1.9\times$ higher throughput and $1.8\times$ more resident requests.
These results show that adaptive KV budgeting better translates limited GPU memory into serving concurrency and throughput under heterogeneous workloads.

\subsection{Analysis and Discussion}
\label{sec:analysis}

\begin{figure*}[t]
    \centering
    \begin{minipage}[t]{0.47\textwidth}
        \centering
        \includegraphics[height=0.16\textheight, width=\linewidth, keepaspectratio]{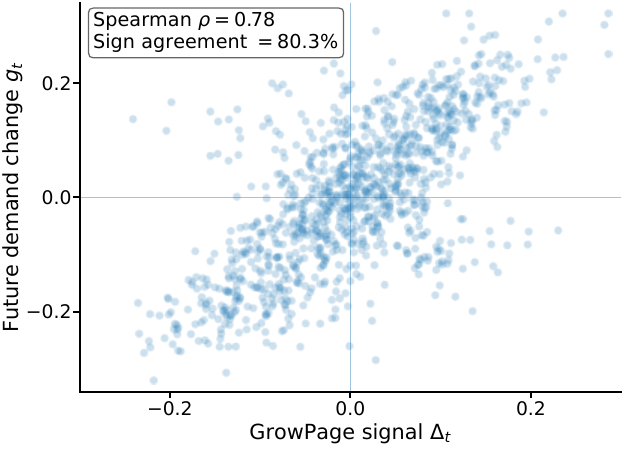}
        \vspace{-4mm}
        \caption{
        Correlation between the GrowPage signal $\Delta_t$ and future attention-demand change on Qwen3-8B with AMC23.
        }
        \label{fig:signal_correlation}
    \end{minipage}
    \hfill
    \begin{minipage}[t]{0.47\textwidth}
        \centering
        \includegraphics[height=0.16\textheight, width=\linewidth, keepaspectratio]{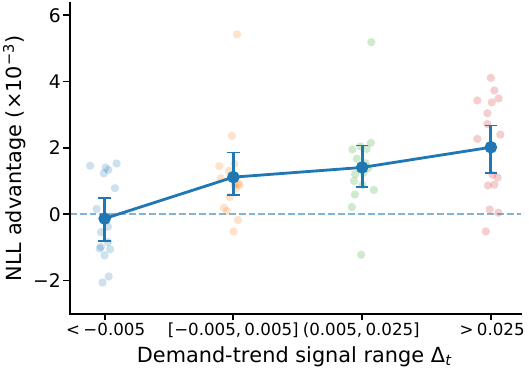}
        \vspace{-4mm}
        \caption{
        Counterfactual validation of GrowPage adaptive capacity decisions on Qwen3-8B with AMC23.
        }
        \label{fig:capacity_validation}
    \end{minipage}
    \vspace{-4mm}
\end{figure*}

\paragraph{Future Demand Evolution Prediction.}
We first evaluate whether $\Delta_t$ reflects the actual change of future KV demand.
Specifically, we define the future demand change as the variation between the current attention working set and the actual attention working set of tokens generated within the next physical KV page.
A positive value indicates that future tokens require a broader historical working set, while a negative value indicates increasing attention concentration.
Figure~\ref{fig:signal_correlation} shows the correlation between $\Delta_t$ and future demand change on Qwen3-8B with AMC23.
The two variables exhibit a strong monotonic relationship, achieving a Spearman correlation coefficient of $\rho=0.78$ and a sign agreement of 80.3\%.
This demonstrates that the dual-timescale query summaries effectively capture the temporal evolution of attention demand.

\begin{wraptable}{R}{0.42\textwidth}
\vspace{-6mm}
\begin{minipage}{\linewidth}
\centering
\caption{{Ablation of dual-timescale demand estimation on Qwen3-8B.}}
\renewcommand*{\arraystretch}{1.0}
\setlength\tabcolsep{3pt}

\resizebox{\linewidth}{!}{
\begin{tabular}{c|c|c|cc|cc}
\Xhline{2\arrayrulewidth}

\multirow{2}{*}{$\beta_S$}
&\multirow{2}{*}{$\beta_L$}
& \multirow{2}{*}{Metric}
& \multicolumn{2}{c|}{AMC23}
& \multicolumn{2}{c}{AIME24}
\\

\cline{4-7}

&
&
& Pass@1
& TPS
& Pass@1
& TPS
\\

\Xhline{2\arrayrulewidth}

0
& 0.999
& Top-$p$
& 89.9
& 2394
& 71.2
& 1264
\\

0
& 0.99
& Top-$p$
& 89.1
& \textbf{2478}
& 69.9
& \textbf{1317}
\\

0.9
& 0.99
& Top-$p$
& 90.3
& 2362
& 71.7
& 1256
\\

0.99
& 0.999
& Top-$p$
& 90.9
& 2308
& 72.9
& 1218
\\

0.9
& 0.999
& Entropy
& 90.5
& 2179
& 72.4
& 1136
\\

\cellcolor[HTML]{D5E8D4}\textbf{0.9}
& \cellcolor[HTML]{D5E8D4}\textbf{0.999}
& \cellcolor[HTML]{D5E8D4}\textbf{Top-$p$}
& \cellcolor[HTML]{D5E8D4}\textbf{91.4}
& \cellcolor[HTML]{D5E8D4}2261
& \cellcolor[HTML]{D5E8D4}\textbf{73.8}
& \cellcolor[HTML]{D5E8D4}1190
\\

\Xhline{2\arrayrulewidth}
\end{tabular}}

\label{tab:signal_ablation}
\end{minipage}
\vspace{-4mm}
\end{wraptable}

\paragraph{Capacity Expansion Benefit Analysis.}
We further investigate whether $\Delta_t$ can identify when allocating additional KV capacity is beneficial.
At each capacity boundary, we construct two counterfactual KV states corresponding to \emph{Compress \& Hold} and \emph{Grow by One Page}.
To avoid the influence of different generation trajectories, we first obtain a reference decoding trace using full KV inference and replay the subsequent tokens under the two counterfactual KV states.
Both branches share the same model parameters and future token sequence from the reference trajectory, differing only in the available KV capacity.
We evaluate the prediction quality of the replayed tokens using teacher-forced negative log-likelihood (NLL).
Let $L_{\mathrm{compress}}$ and $L_{\mathrm{grow}}$ denote the average NLL under \emph{Compress \& Hold} and \emph{Grow by One Page}, respectively.
We define the growth advantage as $A_t=L_{\mathrm{compress}}-L_{\mathrm{grow}}$, where a larger $A_t$ indicates a larger reduction in prediction loss from allocating additional KV capacity.
Figure~\ref{fig:capacity_validation} groups samples according to different $\Delta_t$ ranges and reports the corresponding growth advantage.
The growth advantage consistently increases as $\Delta_t$ becomes larger, demonstrating that the proposed demand-trend signal not only captures attention-demand variation but also provides a reliable indicator for capacity expansion decisions.
\begin{wrapfigure}{R}{0.50\textwidth}
\vspace{-6mm}
  \centering \includegraphics[width = 0.49\textwidth, keepaspectratio]{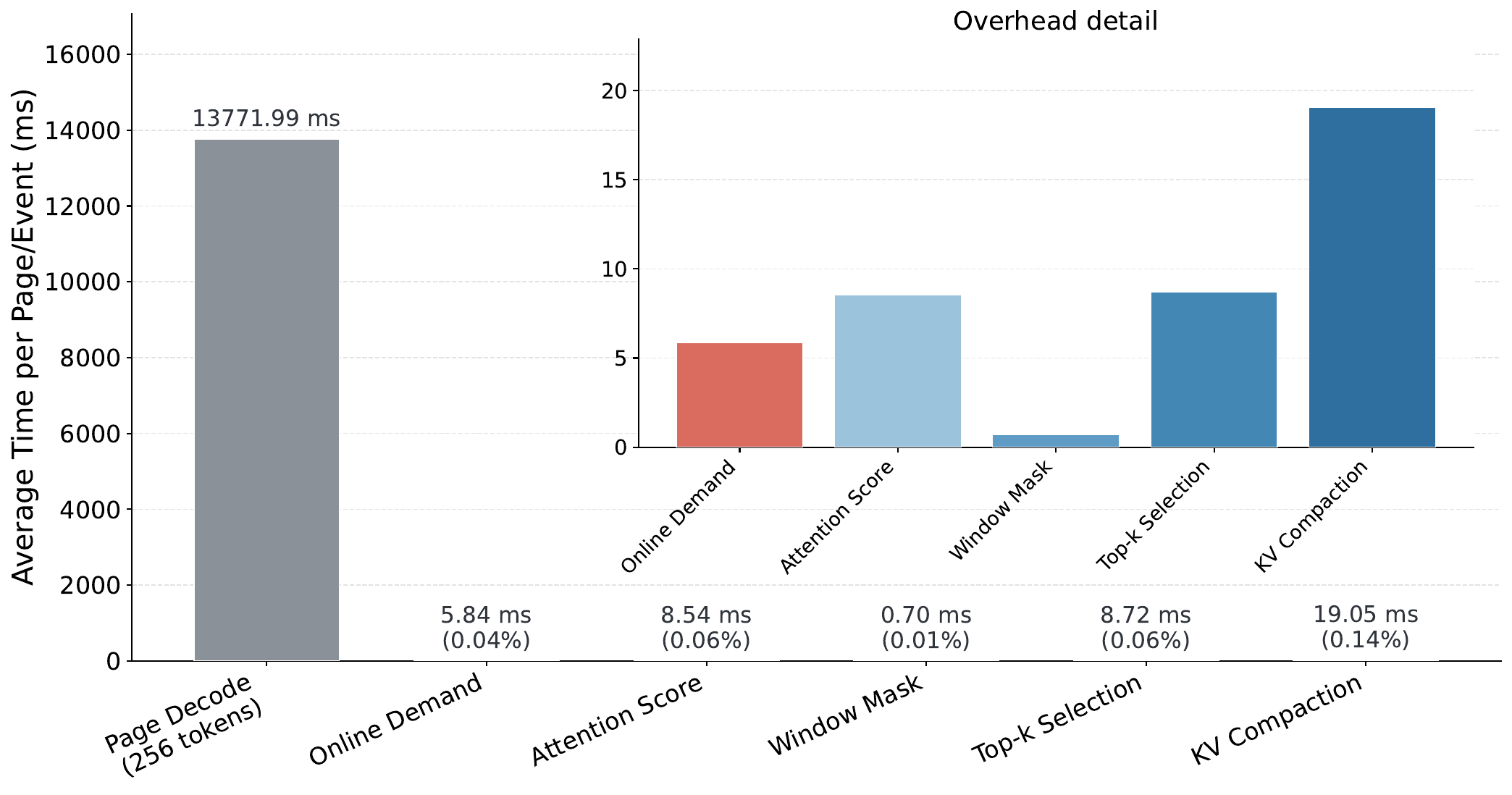}
  \vspace{-2mm}
  \caption{{Runtime overhead analysis of GrowPage on Qwen3-8B with AMC23.}}
  \vspace{-3mm}
  \label{fig:overhead}
\end{wrapfigure}

\paragraph{Dual-Timescale Signal Ablation.}
We further ablate the design of the dual-timescale demand signal on Qwen3-8B, with results reported in Table~\ref{tab:signal_ablation}.
Here, $\beta_S=0$ reduces the short-term summary to the current query.
Across both AMC23 and AIME24, using the current query directly yields higher throughput but noticeably lower reasoning accuracy, while introducing moderate short-term smoothing improves the accuracy--throughput trade-off.
A longer-term reference $\beta_L=0.999$ also consistently improves accuracy over $\beta_L=0.99$ under the same short-term setting.
However, overly smoothing the short-term summary $\beta_S=0.99$ weakens its responsiveness and performs worse than the default $(\beta_S,\beta_L)=(0.9,0.999)$, which achieves the highest pass@1 accuracy of 91.4\% on AMC23 and 73.8\% on AIME24.
Replacing the Top-$p$ working-set metric with attention entropy further degrades both accuracy and throughput under the same EMA configuration.
These results support the use of well-separated temporal summaries together with the Top-$p$ working set for robust online demand estimation.

\paragraph{Runtime Overhead Analysis.}
GrowPage invokes demand estimation and capacity control only at capacity boundaries rather than at every decoding step.
Figure~\ref{fig:overhead} shows that Online Demand, Attention Score, Window Mask, Top-k Selection, and KV Compaction incur only 5.84, 8.54, 0.70, 8.72, and 19.05 ms per event, respectively, accounting for less than 0.32\% of the corresponding decoding time.
KV Compaction dominates the additional overhead, while the overall runtime cost remains negligible relative to long-output decoding.

\section{Related Work}

\paragraph{KV-cache compression for LLM inference.}
KV-cache compression has become an important direction for reducing the memory footprint of long-context and reasoning inference.
Early approaches mainly focus on retaining important tokens under a limited cache budget, such as StreamingLLM~\citep{xiao2024efficient}, SnapKV~\citep{li2024snapkv}, and PyramidKV~\citep{cai2024pyramidkv}.
Recent studies extend KV eviction to the decoding stage to control cache growth during long-form generation.
Representative methods include H2O~\citep{zhang2023h2o}, Quest~\citep{tang2024quest}, R-KV~\citep{cai2026r}, G-KV~\citep{liao2025g}, and ThinkKV~\citep{ramachandran2026thinkv}.
These methods mainly improve token retention or eviction policies under a predefined KV budget.
However, the allocated capacity itself is typically fixed during decoding, which limits their ability to adapt to heterogeneous and evolving KV demands across reasoning requests.

\paragraph{System-aware KV optimization.}
Beyond token selection algorithms, recent works explore integrating KV management with practical LLM serving systems.
PagedAttention~\citep{kwon2023efficient} introduces a page-based memory abstraction that enables efficient KV management under continuous batching.
Recent system-level KV management approaches further investigate efficient page organization and KV scheduling for long-sequence inference~\citep{maoefficient}.
Zipage~\citep{liao2026zipage} integrates token-level KV eviction with PagedAttention through Compressed PagedAttention, preserving important serving features such as prefix caching.
Nevertheless, existing system-compatible approaches still assume a predefined per-request KV capacity.
In contrast, GrowPage treats KV capacity itself as a runtime resource and dynamically maps demand estimation into page-level allocation decisions.

\section{Conclusion}
Large reasoning models introduce substantial KV-cache pressure due to their long chain-of-thought generation, making efficient memory management critical for LLM reasoning serving.
Existing KV compression methods typically rely on fixed per-request capacity budgets, which cannot accommodate the heterogeneous and evolving KV demands of reasoning requests.
In this work, we present GrowPage, an on-demand KV budgeting framework that treats KV capacity as a runtime resource.
GrowPage uses lightweight dual-timescale query summaries to estimate demand evolution and dynamically chooses between KV compaction within the current allocation and incremental physical-page expansion.
By integrating this capability with PagedAttention-based serving, GrowPage preserves key system optimizations including continuous batching and prefix caching while achieving improved reasoning performance--throughput trade-offs across diverse benchmarks.

\bibliography{iclr2027_conference}
\bibliographystyle{iclr2027_conference}

\appendix
\startcontents[appendix]

\printcontents[appendix]{l}{1}{%
  \section*{\textbf{\huge{Appendix}}}
  \setcounter{tocdepth}{2} 
}
\newpage
\section{Overview of Mathematical Notation}
\label{sec:appendix_notations}

For clarity, Table~\ref{tab:notation} summarizes the key notation used in the theoretical analysis and the GrowPage capacity-control framework.

\begingroup
\begin{table}[t]
\centering
\caption{Summary of key notation used in the paper.}
\renewcommand*{\arraystretch}{1.05}
\setlength\tabcolsep{5pt}

\resizebox{0.88\linewidth}{!}{%
\begin{tabular}{ll}
\toprule
\textbf{Symbol} & \textbf{Description} \\
\midrule

$B_i^{\star}$
& Minimum sufficient KV budget of request $i$ \\

$\mathbf{a}_t$
& Attention distribution over historical KV states at decoding step $t$ \\

$D_p(\mathbf{a}_t)$
& Minimum number of KV states required to cover attention mass $p$ \\

$S$
& Subset of retained historical KV states \\

$m_t(S)$
& Attention mass retained by subset $S$ \\

$\mathbf{o}_t,\widehat{\mathbf{o}}_t(S)$
& Full-cache and subset-renormalized attention outputs \\

$C_{\delta}^{\mathrm{wc}}(\mathbf{a}_t)$
& Minimum worst-case KV capacity under approximation tolerance $\delta$ \\

$\mathbf{q}_t$
& Current normalized pre-RoPE query \\

$\bar{\mathbf{q}}_t^{S},\bar{\mathbf{q}}_t^{L}$
& Short- and long-timescale EMA query summaries \\

$\beta_S,\beta_L$
& EMA decay factors for the two temporal query summaries \\

$\mathcal{C}_t$
& Historical candidate KV set for online demand estimation \\

$K_{t,\ell,h}^{x}$
& Top-$p$ working-set size for temporal view $x\in\{S,L\}$ \\

$r_t^{S},r_t^{L}$
& Request-level short- and long-timescale working-set ratios \\

$\Delta_t$
& Demand-trend signal, defined as $r_t^{S}-r_t^{L}$ \\

$\tau$
& Threshold controlling compression versus capacity growth \\

$s_{i,\ell,h}$
& KV importance score for token $i$ at layer $\ell$ and head $h$ \\

$P_t$
& Number of physical KV pages allocated to the request \\

$b$
& Number of KV-token slots in one physical page \\

$\mathcal{R}_t$
& Protected recent-token set excluded from compression \\

$K_t$
& Historical KV retention budget at a compression event \\

$k_{\mathrm{loc}}$
& Block-local KV retention quota \\

\bottomrule
\end{tabular}}
\vspace{-6pt}

\label{tab:notation}
\end{table}
\endgroup

\section{Extended Related Works}
\label{sec:appendix_related_works}

\paragraph{Fine-Grained KV Compression.} Beyond the representative KV-eviction methods discussed in the main text, a broad body of work explores finer-grained cache adaptation and alternative compression dimensions.
Scissorhands~\citep{liu2023scissorhands} exploits the persistence of token importance to probabilistically retain historically influential KV states, while Keyformer~\citep{adnan2024keyformer} identifies a compact set of key tokens that dominate attention during generation.
Rather than applying a uniform policy to all attention heads, FastGen~\citep{ge2024model} profiles their attention structures and assigns different cache strategies accordingly, and DuoAttention~\citep{xiao2025duoattention} separates retrieval heads requiring long-context access from streaming heads that can operate with compact caches.
NACL~\citep{chen2024nacl} combines proxy-token-based importance with randomized eviction, whereas BUZZ~\citep{zhao2026buzz} preserves recent context together with segmented historical heavy hitters.
A complementary direction reduces the representation cost of retained KV states.
KIVI~\citep{liu2024kivi} introduces asymmetric low-bit quantization tailored to the distinct distributions of keys and values, and KVQuant~\citep{hooper2024kvquant} combines pre-RoPE key quantization, non-uniform datatypes, and explicit outlier handling.
MiKV~\citep{yang2024no} preserves important KV states at higher precision while representing less important states at lower precision, and GEAR~\citep{kang2024gear} combines quantization with low-rank and sparse error correction.
Cross-layer redundancy has also been exploited by MiniCache~\citep{liu2024minicache}, which merges similar KV states across neighboring layers.
These approaches adapt \emph{which} states are represented and \emph{how} they are represented, but do not directly address online request-level decisions about how much total KV capacity should be allocated as generation evolves.

\paragraph{KV Compression for Long Reasoning.} Recent work has increasingly specialized KV-cache compression for long-output reasoning, where token importance can change substantially over an extended generation trajectory.
LazyEviction~\citep{zhang2026lazyeviction} identifies \emph{token importance recurrence}, observing that previously unimportant tokens may regain high attention after many decoding steps, and therefore delays irreversible eviction through an observation window.
DesireKV~\citep{cheng2026desirekv} jointly considers attention-derived importance and quantization sensitivity, selectively retaining, quantizing, or evicting reasoning tokens according to their different roles.
LongFlow~\citep{su2026longflow} derives token-importance estimates directly from intermediate attention computations and co-designs them with fused kernels to reduce the overhead of continuous importance evaluation during long reasoning.
InfoKV~\citep{kai2026information} further argues that attention alone may overlook tokens with long-range future influence and supplements attention scores with information-theoretic signals related to predictive uncertainty.
Kara~\citep{han2026kara} restricts compression to a sliding window of recently generated reasoning context and develops a serving implementation compatible with paged KV management.
REAL~\citep{li2026real} instead analyzes heterogeneous attention-head behaviors in both successful and failed retrieval-reasoning cases to construct more robust eviction policies.
Together, these studies reinforce that reasoning-time KV importance is highly dynamic.
Nevertheless, their adaptation primarily concerns importance estimation, eviction timing, token grouping, or representation precision; GrowPage operates on an orthogonal dimension by adapting the \emph{total per-request capacity} according to the evolving demand of the request.

\paragraph{System-Level KV Management.}
Another related direction optimizes where and how KV states are physically stored and transferred in serving systems.
vAttention~\citep{prabhu2025vattention} uses CUDA virtual memory to support dynamic physical-memory allocation while retaining a contiguous virtual KV layout, illustrating that the physical storage backing an active sequence can be managed independently from its logical address space.
For memory hierarchies extending beyond GPU memory, InfiniGen~\citep{lee2024infinigen} predicts important KV entries and selectively fetches them from CPU memory, thereby reducing the transfer overhead of conventional KV offloading.
CacheGen~\citep{liu2024cachegen} compresses reusable KV caches into compact bitstreams for efficient context loading and adapts compression to network conditions.
At cluster scale, MemServe~\citep{hu2024memserve} introduces an elastic distributed memory pool for context caching and disaggregated inference, while Mooncake~\citep{qin2026mooncake} builds a KV-cache-centric serving architecture that exploits heterogeneous CPU, DRAM, SSD, and network resources together with a global scheduler.
These systems mainly improve KV placement, movement, reuse, or physical memory provisioning.
GrowPage addresses a different but complementary problem: before allocating additional GPU-resident KV storage to an active reasoning request, it estimates whether the request's evolving attention demand justifies expanding its logical cache capacity.
This demand-aware capacity control allows the page-based memory manager to respond not only to sequence growth, but also to heterogeneous and time-varying information requirements.

\section{Proof of Theorem~\ref{thm:attention_capacity}}
\label{app:attention_capacity_proof}

For a retained subset $S$, let
\[
m(S)=\sum_{i\in S}a_{t,i}.
\]
The difference between the full and renormalized attention outputs is
\[
\mathbf{o}_t-\widehat{\mathbf{o}}_t(S)
=
\sum_{i\notin S}a_{t,i}\mathbf{v}_i
-
\frac{1-m(S)}{m(S)}
\sum_{i\in S}a_{t,i}\mathbf{v}_i .
\]
Using $\|\mathbf{v}_i\|_2\le V_{\max}$ and the triangle inequality,
\[
\begin{aligned}
\left\|
\mathbf{o}_t-\widehat{\mathbf{o}}_t(S)
\right\|_2
&\le
\sum_{i\notin S}a_{t,i}V_{\max}
+
\frac{1-m(S)}{m(S)}
\sum_{i\in S}a_{t,i}V_{\max}\\
&=
2V_{\max}(1-m(S)).
\end{aligned}
\]

This bound is tight. Let $\mathbf{u}$ be any unit vector and choose
$\mathbf{v}_i=-V_{\max}\mathbf{u}$ for $i\in S$ and
$\mathbf{v}_i=V_{\max}\mathbf{u}$ for $i\notin S$.
Then
\[
\left\|
\mathbf{o}_t-\widehat{\mathbf{o}}_t(S)
\right\|_2
=
2V_{\max}(1-m(S)).
\]
Hence,
\[
\sup_{\|\mathbf{v}_i\|_2\le V_{\max}}
\left\|
\mathbf{o}_t-\widehat{\mathbf{o}}_t(S)
\right\|_2
=
2V_{\max}(1-m(S)).
\]
Therefore, guaranteeing error at most $\delta$ for all admissible value vectors is equivalent to
\[
m(S)\ge 1-\frac{\delta}{2V_{\max}}.
\]
Among all subsets of cardinality $k$, the maximum retained attention mass is obtained by selecting the $k$ largest attention weights.
Thus, the minimum cardinality satisfying the above condition is
\[
C_\delta^{\mathrm{wc}}(\mathbf{a}_t)
=
D_{1-\delta/(2V_{\max})}(\mathbf{a}_t),
\]
which proves the theorem.

\begin{wrapfigure}{R}{0.50\textwidth}
\vspace{-6mm}
  \centering \includegraphics[width = 0.49\textwidth, keepaspectratio]{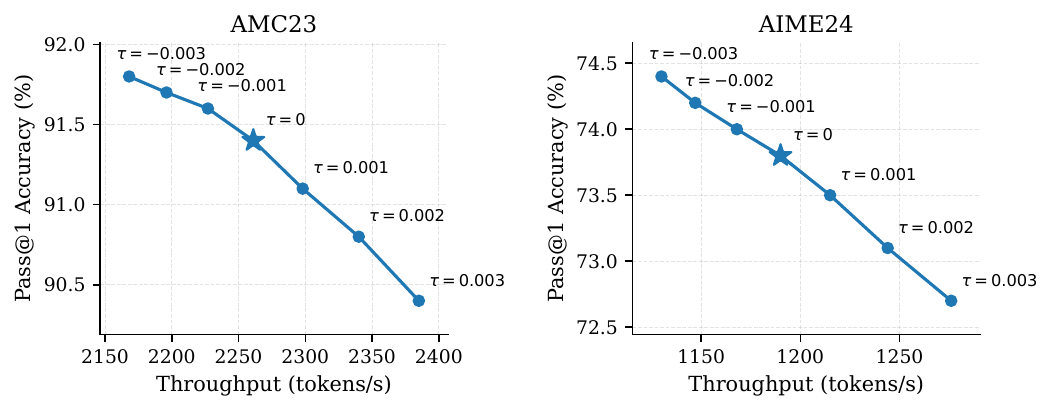}
  \vspace{-2mm}
  \caption{
Accuracy--throughput trade-offs under different demand thresholds $\tau$ on Qwen3-8B with AMC23 and AIME24.
}
\label{fig:tau_sensitivity}
\end{wrapfigure}

\section{Hyperparameter Sensitivity}
\label{app:hyperparameter}

\subsection{Effect of Demand Threshold}
\label{app:demand_threshold}

We further study the sensitivity of GrowPage to the demand threshold $\tau$, which controls the trade-off between KV compression and capacity expansion.
As shown by the counterfactual analysis in Figure~\ref{fig:capacity_validation}, the benefit of allocating additional KV capacity changes noticeably as $\Delta_t$ crosses the region around zero.
We therefore focus on the fine-grained interval around $[-0.005,0.005]$ and vary $\tau$ from $-0.003$ to $0.003$ with a step size of $0.001$.
All experiments are conducted on Qwen3-8B using AMC23 and AIME24.
Figure~\ref{fig:tau_sensitivity} shows that $\tau$ provides a direct mechanism for controlling the accuracy--throughput operating point of GrowPage.
A smaller threshold makes capacity expansion more permissive, causing more requests to acquire additional KV pages.
This preserves a larger historical working set and generally improves reasoning accuracy, but increases per-request KV consumption and consequently reduces serving throughput.
Conversely, a larger $\tau$ makes GrowPage more conservative about expansion, resulting in more frequent \emph{Compress \& Hold} decisions and higher throughput at the cost of more aggressive information removal.
Across both AMC23 and AIME24, $\tau=0$ provides a stable operating point with a balanced trade-off between reasoning accuracy and serving throughput.
We therefore adopt $\tau=0$ as the default threshold throughout our experiments.
\begin{wraptable}{R}{0.29\textwidth}
\vspace{-5mm}
\begin{minipage}{\linewidth}
\centering
\caption{Effect of the local retention quota $k_{\mathrm{loc}}$ on Qwen3-8B.}
\renewcommand*{\arraystretch}{1.0}
\setlength\tabcolsep{6pt}

\resizebox{\linewidth}{!}{
\begin{tabular}{c|cc}
\Xhline{2\arrayrulewidth}

\multirow{2}{*}{$k_{\mathrm{loc}}$}
& AMC23
& AIME24
\\

&
Pass@1
& Pass@1
\\

\Xhline{2\arrayrulewidth}

16
& 90.3
& 72.4
\\

32
& 90.9
& 73.1
\\

\cellcolor[HTML]{D5E8D4}\textbf{64}
& \cellcolor[HTML]{D5E8D4}\textbf{91.4}
& \cellcolor[HTML]{D5E8D4}\textbf{73.8}
\\

128
& 90.6
& 72.8
\\

\Xhline{2\arrayrulewidth}
\end{tabular}}

\label{tab:local_retention_ablation}
\end{minipage}
\vspace{-4mm}
\end{wraptable}

\subsection{Effect of Local Retention Quota}
\label{app:local_retention}

As described in Section~\ref{sec:capacity_control}, $k_{\mathrm{loc}}$ specifies the minimum number of tokens retained from each historical block before the remaining budget is filled by global importance-based selection.
We evaluate its sensitivity on Qwen3-8B using AMC23 and AIME24.
We report only Pass@1 because different $k_{\mathrm{loc}}$ values yield nearly identical TPS; $k_{\mathrm{loc}}$ mainly affects which KV states are retained rather than the total retained capacity, and therefore primarily influences reasoning accuracy.
As shown in Table~\ref{tab:local_retention_ablation}, performance first improves as $k_{\mathrm{loc}}$ increases from 16 to 64, indicating that sufficient block-local coverage helps prevent globally important regions from dominating the retained KV states.
However, further increasing $k_{\mathrm{loc}}$ to 128 slightly degrades accuracy, since an overly large local quota leaves less flexibility for global selection and forces more tokens to be preserved from relatively unimportant blocks.
Overall, $k_{\mathrm{loc}}=64$ achieves the best performance on both benchmarks, and we therefore use it as the default setting throughout our experiments.

\section{System Implementation Details}
\label{app:system_implementation}

This section provides additional implementation details for the system integration described in Section~\ref{sec:system}.
We first describe the runtime management and computation of the dual-timescale query states.
Our support for prefix caching and asynchronous KV compression largely follows the system design of Zipage~\citep{liao2026zipage}, with the compression target adapted to GrowPage's dynamically determined page capacity.

\subsection{Dual-Timescale Query Management and Online Demand Estimation}
\label{app:dual_query_implementation}

\paragraph{Pre-allocated query-state pool.}
To avoid request-level GPU allocation and deallocation, GrowPage pre-allocates a fixed query-state tensor when the inference engine is initialized.
Its layout is
\begin{equation}
\mathcal{Q}
\in
\mathbb{R}^{
L\times M\times 2\times H_q\times d_h},
\label{eq:query_cache_layout}
\end{equation}
where $L$ is the number of attention layers, $M$ is the maximum number of query slots, $H_q$ is the number of query heads per GPU, and $d_h$ is the head dimension.
The third dimension stores the short- and long-timescale EMA queries.
We set $M$ to the serving engine's maximum number of concurrent sequences (\texttt{max\_num\_seqs}), which is 512 by default.

Query slots are bound to requests only when they enter the running queue.
The scheduler maintains a pool of free slot identifiers and assigns one identifier to each newly admitted running request; requests waiting to be scheduled therefore consume no query-state slot.
When a request finishes, its identifier is returned to the free pool without releasing the underlying GPU tensor.
The same procedure is applied when a request is preempted; if it is scheduled again, it acquires a new slot and reinitializes its summaries.
This slot-based design avoids frequent CUDA memory allocation and allows the pre-allocated storage to be continuously reused across requests.
The resulting auxiliary memory is fixed with respect to sequence length and scales as
$2MLH_qd_h$ elements.

\paragraph{Initialization and online update.}
The query summaries are updated only during decoding.
For the first decoding step of a newly assigned slot, both temporal states are initialized directly from the current normalized pre-RoPE query,
$\bar{\mathbf q}^{S}=\bar{\mathbf q}^{L}=\mathbf q_t$,
thereby overwriting any stale values left by the previous owner of the slot without explicitly clearing the tensor.
Subsequent decoding steps update the two states in place according to Eq.~\ref{eq:dual_ema}, using $\beta_S=0.9$ and $\beta_L=0.999$.
No historical query sequence is materialized: each request stores only the two running summaries.

\paragraph{Demand estimation at capacity boundaries.}
The regular decoding path only updates the two EMA states.
The remaining demand-estimation operations are executed when a request reaches a capacity boundary.
Because EMA averaging changes the magnitude of the stored queries, we first calibrate each temporal summary to the RMS scale of the current normalized query.
For $x\in\{S,L\}$, we compute
\begin{equation}
\gamma_t^{x}
=
\frac{\operatorname{RMS}(\mathbf q_t)}
     {\operatorname{RMS}(\bar{\mathbf q}_t^{x})},
\qquad
\widetilde{\mathbf q}_t^{x}
=
\gamma_t^{x}\bar{\mathbf q}_t^{x}.
\label{eq:rms_calibration}
\end{equation}
The calibrated queries are then transformed by RoPE using the \emph{current} decoding position.
This preserves the temporal information accumulated in the common pre-RoPE space while aligning both summaries with the positional frame used by the current attention computation.

We exclude the most recent physical KV block from the historical candidate set, corresponding to 256 tokens under our default block size, and independently compute the short- and long-timescale attention distributions over the remaining historical keys.
The native GQA head mapping is preserved when matching query heads to KV heads.
For each temporal view, the attention weights are sorted and accumulated until reaching the Top-$p$ coverage threshold ($p=0.99$), yielding the working-set ratios in Eq.~\ref{eq:working_set_ratio}.
These ratios are then aggregated across layers and KV heads according to Eq.~\ref{eq:demand_aggregation}, producing the request-level signal $\Delta_t$.
Finally, $\Delta_t$ is compared with the demand threshold $\tau$ to select between \emph{Compress \& Hold} and \emph{Grow by One Page}.
Thus, attention scoring, sorting, and working-set estimation are kept off the critical path of ordinary decoding steps and are invoked only when a capacity decision is required.

\subsection{Prefix-Caching Support}
\label{app:prefix_caching}

GrowPage inherits the prefix-preserving compression mechanism of Zipage~\citep{liao2026zipage}.
Prefix sharing is tracked at physical-block granularity using reference counts, and a block referenced by multiple requests is never overwritten in place during compression.
Instead, compression writes the retained KV states into a set of target blocks while leaving shared prefix blocks intact for the other requests.
We adapt the target size to GrowPage's dynamic capacity.
For a request currently occupying $P_t$ physical pages, \emph{Compress \& Hold} keeps the physical allocation unchanged but compacts the retained KV states into at most $(P_t-1)b$ occupied slots, leaving one page worth of free slots for subsequent decoding.
If shared prefix blocks occupy part of the original allocation, sufficient new target blocks are allocated to replace those shared blocks, while reusable non-shared blocks are retained whenever possible.
After compaction, the request releases its references to the original shared blocks and their reference counts are updated accordingly.
\emph{Grow by One Page} does not modify the shared prefix and simply appends the newly allocated page to the request's block table.
As in Zipage, temporary target-block allocation may precede the release of shared blocks; under severe memory pressure, the scheduler can therefore invoke preemption to avoid allocation deadlock.

\subsection{Asynchronous Decoding and Compression}
\label{app:asynchronous_compression}

We also adopt the asynchronous decoding--compression pipeline of Zipage~\citep{liao2026zipage}.
At a given scheduling step, only a subset of requests reaching capacity boundaries require KV compaction.
Executing compression synchronously would stall the remaining decode-ready requests and serialize a relatively small compression batch with the main decoding workload.
Instead, requests selected for \emph{Compress \& Hold} enter the asynchronous compression path, while other ready requests continue decoding.
Once compaction completes, the compressed requests rejoin a subsequent decoding batch.
Requests taking \emph{Grow by One Page} require only page allocation and block-table update and can resume decoding after the new page is installed.
This preserves Zipage's overlap between decoding and KV compaction while allowing GrowPage's capacity decisions to remain request-specific.

\section{Experimental Details}
\subsection{Datasets and Evaluation Protocol}
\label{app:dataset_details}
We evaluate GrowPage on five reasoning benchmarks covering mathematical reasoning and code generation.
Table~\ref{tab:dataset_settings} summarizes the dataset size, number of independent generations per question, maximum output length, and the KV budget used by fixed-budget compression baselines.
Unless otherwise specified, we use a sampling temperature of 0.6.
For benchmarks with multiple generations per question, each generation is evaluated independently and the reported Pass@1 is averaged over all sampled outputs.
The dataset-specific KV budgets follow the configurations used in our fixed-budget comparisons; GrowPage does not use these predefined budgets and instead adapts its KV capacity online.

\begin{table}[t]
\centering
\caption{Evaluation settings for individual reasoning benchmarks. 
The KV budget denotes the per-request capacity used by fixed-budget compression baselines.}
\label{tab:dataset_settings}

\renewcommand*{\arraystretch}{1.05}
\setlength\tabcolsep{6pt}

\resizebox{0.88\linewidth}{!}{
\begin{tabular}{lcccc}
\Xhline{2\arrayrulewidth}

\textbf{Workload}
& \textbf{Number of Questions}
& \textbf{Sample Times}
& \textbf{Max Output Length}
& \textbf{KV Budget}
\\

\Xhline{2\arrayrulewidth}

AMC23
& 40
& 32
& 16384
& 4096
\\

AIME24
& 30
& 32
& 32768
& 4096
\\

MATH500
& 500
& 8
& 16384
& 4096
\\

GSM8K
& 1319
& 4
& 16384
& 4096
\\

LiveCodeBench v1
& 400
& 8
& 16384
& 8192
\\

\Xhline{2\arrayrulewidth}
\end{tabular}
}
\end{table}
The sampling multiplicity is chosen according to the size and difficulty of each benchmark.
For the smaller competition-level mathematics datasets AMC23 and AIME24, we use 32 independent generations per question to reduce sampling variance.
For the larger MATH500, GSM8K, and LiveCodeBench datasets, fewer repetitions provide sufficient coverage while keeping the total evaluation cost manageable.
The larger KV budget for LiveCodeBench reflects its substantially longer code-generation trajectories.

\subsection{Mixed-Workload Construction}
\label{app:mixed_workload}

For the heterogeneous-workload experiment in Figure~\ref{fig:mixed_pareto}, we combine AMC23, AIME24, and GSM8K into a single serving workload.
Because these datasets differ substantially in the number of available questions, directly using each question once would cause GSM8K to dominate the workload.
We therefore use different sampling multiplicities to obtain a more balanced contribution from the three datasets.
The resulting workload contains 3559 requests, as summarized in Table~\ref{tab:mixed_workload}.
All requests use a maximum output length of 32768 tokens.

\begin{wraptable}{R}{0.38\textwidth}
\vspace{-5mm}
\begin{minipage}{\linewidth}
\centering
\caption{Composition of the mixed reasoning workload.}
\renewcommand*{\arraystretch}{1.0}
\setlength\tabcolsep{5pt}

\resizebox{\linewidth}{!}{
\begin{tabular}{l|ccc}
\Xhline{2\arrayrulewidth}

\textbf{Dataset}
& \textbf{Number of Questions}
& \textbf{Samples}
& \textbf{Total Requests}
\\

\Xhline{2\arrayrulewidth}

AMC23
& 40
& 32
& 1280
\\

AIME24
& 30
& 32
& 960
\\

GSM8K
& 1319
& 1
& 1319
\\

\Xhline{2\arrayrulewidth}
\end{tabular}}

\label{tab:mixed_workload}
\end{minipage}
\vspace{-4mm}
\end{wraptable}

This construction prevents the much larger GSM8K dataset from overwhelmingly dominating the mixed workload and yields comparable numbers of requests from the three constituent benchmarks.
Unlike the individual benchmark comparison in Table~\ref{tab:dataset_settings}, we do not assign a single fixed KV budget to this workload, since the purpose of this experiment is to explicitly evaluate Zipage across multiple fixed budgets and compare its resulting accuracy--throughput Pareto frontier with GrowPage.

\section{Extended Analysis of Main Results}
\label{app:main_results_analysis}

\subsection{Additional Mixed-Workload Results}
\label{app:mixed_results_llama}

To examine whether the mixed-workload behavior in Figure~\ref{fig:mixed_pareto} generalizes across model architectures, we repeat the same experiment on DeepSeek-R1-Distill-Llama-8B.
We use the identical mixed workload described in Appendix~\ref{app:mixed_workload}, and compare GrowPage with FullKV and Zipage under fixed KV budgets ranging from 1024 to 8192 tokens.

\begin{wrapfigure}{R}{0.40\textwidth}
\vspace{-6mm}
  \centering \includegraphics[width = 0.39\textwidth, keepaspectratio]{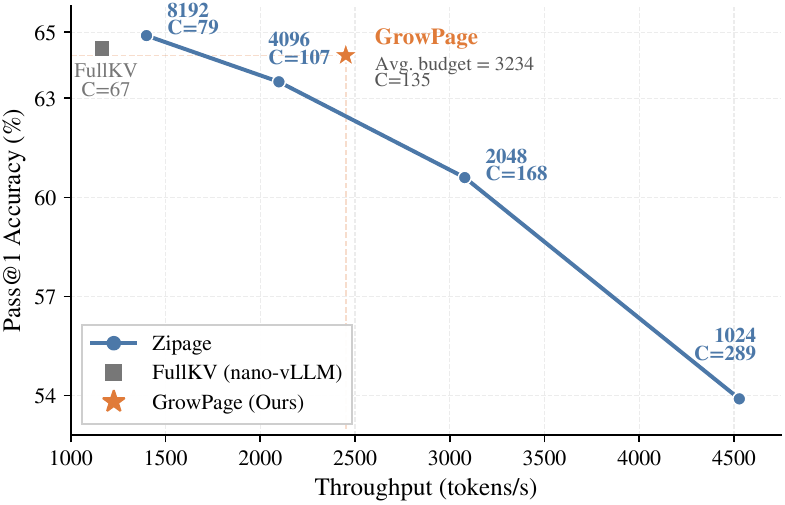}
  \vspace{-4mm}
  \caption{
Accuracy--throughput trade-off on the mixed GSM8K, AMC23, and AIME24 workload with DeepSeek-R1-Distill-Llama-8B. $C$ denotes the average number of requests resident in GPU memory.
}
  \vspace{-3mm}
  \label{fig:mixed_pareto_llama}
\end{wrapfigure}

As shown in Figure~\ref{fig:mixed_pareto_llama}, DeepSeek-R1-Distill-Llama-8B exhibits a trend consistent with the Qwen3-8B results in Figure~\ref{fig:mixed_pareto}.
Increasing the fixed KV budget of Zipage improves reasoning accuracy but progressively reduces resident concurrency and serving throughput.
GrowPage instead reaches 64.3\% Pass@1 at 2451 tokens/s with an average KV budget of 3234 tokens and $C=135$.
Compared with Zipage-4096, GrowPage improves accuracy from 63.5\% to 64.3\% while increasing throughput from 2098 to 2451 tokens/s and resident concurrency from 107 to 135.
Compared with the more conservative Zipage-8192 setting, GrowPage remains within 0.6 percentage points in accuracy while achieving $1.75\times$ higher throughput and $1.71\times$ more resident requests.
It also closely matches FullKV accuracy (64.3\% vs.\ 64.5\%) while providing $2.11\times$ higher throughput.
These results further show that on-demand KV budgeting achieves a favorable accuracy--efficiency trade-off across different reasoning-model architectures rather than being specific to Qwen3-8B.

\subsection{Detailed Serving Metrics}
\label{app:detailed_serving_metrics}

Table~\ref{tab:extended_main_results} provides additional efficiency metrics for the main comparison in Table~\ref{tab:main_results}.
We report the total inference time for each benchmark and the time per output token (TPOT).
The total time is converted from seconds to hours for readability.
Since different benchmarks contain different numbers of sampled requests, total inference time should be compared across methods within the same benchmark rather than across datasets.

\begin{table*}[t]
\centering
\caption{
Extended efficiency comparison on reasoning benchmarks.
We report total inference time (hours) and time per output token (TPOT, ms/token); lower is better for both metrics.
}
\label{tab:extended_main_results}

\setlength{\tabcolsep}{4pt}
\small

\resizebox{1.0\linewidth}{!}{
\begin{tabular}{lcccccccccccc}
\toprule
& \multicolumn{6}{c}{Total Inference Time (h) $\downarrow$}
& \multicolumn{6}{c}{TPOT (ms/token) $\downarrow$}
\\
\cmidrule(lr){2-7}
\cmidrule(lr){8-13}

Method
& GSM8K
& MATH500
& AMC23
& AIME24
& LiveCode
& Avg.
& GSM8K
& MATH500
& AMC23
& AIME24
& LiveCode
& Avg.
\\
\midrule

\multicolumn{13}{l}{\textit{DeepSeek-R1-Distill-Llama-8B}}
\\
\midrule

FullKV (vLLM)
& 0.61 & 3.30 & 2.00 & 5.27 & 8.10 & 3.86
& 117 & 522 & 994 & 783 & 579 & 599.0
\\

FullKV (nano-vLLM)
& 0.64 & 3.44 & 2.06 & 5.25 & 8.51 & 3.98
& 123 & 546 & 1027 & 814 & 598 & 621.6
\\

MorphKV \texttt{{\scriptsize (ICML'25)}}
& 3.95 & 25.99 & 16.09 & 44.49 & 66.13 & 31.33
& 136 & 157 & 156 & 241 & 248 & 187.6
\\

R-KV \texttt{{\scriptsize (NeurIPS'25)}}
& 4.25 & 25.34 & 15.52 & 40.77 & 64.38 & 30.05
& 151 & 169 & 167 & 254 & 272 & 202.6
\\

G-KV \texttt{{\scriptsize (arXiv'25)}}
& 4.10 & 24.59 & 14.78 & 37.33 & 60.81 & 28.32
& 145 & 163 & 162 & 247 & 260 & 195.4
\\

Zipage \texttt{{\scriptsize (ACL'26)}}
& 0.55 & 2.63 & 1.23 & 3.48 & 7.75 & 3.13
& 81 & 128 & 127 & 240 & 225 & 160.2
\\

\cellcolor[HTML]{D5E8D4}\textbf{GrowPage (Ours)}
& \cellcolor[HTML]{D5E8D4}\textbf{0.49}
& \cellcolor[HTML]{D5E8D4}\textbf{2.18}
& \cellcolor[HTML]{D5E8D4}\textbf{1.19}
& \cellcolor[HTML]{D5E8D4}\textbf{2.93}
& \cellcolor[HTML]{D5E8D4}\textbf{5.75}
& \cellcolor[HTML]{D5E8D4}\textbf{2.51}
& \cellcolor[HTML]{D5E8D4}\textbf{78}
& \cellcolor[HTML]{D5E8D4}\textbf{110}
& \cellcolor[HTML]{D5E8D4}\textbf{112}
& \cellcolor[HTML]{D5E8D4}\textbf{207}
& \cellcolor[HTML]{D5E8D4}\textbf{155}
& \cellcolor[HTML]{D5E8D4}\textbf{132.4}
\\

\midrule

\multicolumn{13}{l}{\textit{Qwen3-8B}}
\\
\midrule

FullKV (vLLM)
& 1.23 & 4.09 & 2.66 & 6.58 & 6.35 & 4.18
& 139 & 381 & 826 & 938 & 496 & 556.0
\\

FullKV (nano-vLLM)
& 1.32 & 4.33 & 2.82 & 7.01 & 6.73 & 4.44
& 148 & 404 & 872 & 1003 & 525 & 590.4
\\

MorphKV \texttt{{\scriptsize (ICML'25)}}
& 8.29 & 17.08 & 10.76 & 35.64 & 26.69 & 19.69
& 170 & 165 & 169 & 250 & 232 & 197.2
\\

R-KV \texttt{{\scriptsize (NeurIPS'25)}}
& 8.41 & 16.63 & 10.13 & 30.94 & 25.68 & 18.36
& 176 & 181 & 181 & 263 & 254 & 211.0
\\

G-KV \texttt{{\scriptsize (arXiv'25)}}
& 8.63 & 16.01 & 10.23 & 31.25 & 24.33 & 18.09
& 181 & 174 & 176 & 256 & 246 & 206.6
\\

Zipage \texttt{{\scriptsize (ACL'26)}}
& 1.06 & 3.11 & 1.63 & 4.19 & 5.39 & 3.08
& 79 & 141 & 138 & 249 & 177 & 156.8
\\

\cellcolor[HTML]{D5E8D4}\textbf{GrowPage (Ours)}
& \cellcolor[HTML]{D5E8D4}\textbf{1.04}
& \cellcolor[HTML]{D5E8D4}\textbf{2.60}
& \cellcolor[HTML]{D5E8D4}\textbf{1.42}
& \cellcolor[HTML]{D5E8D4}\textbf{3.77}
& \cellcolor[HTML]{D5E8D4}\textbf{4.67}
& \cellcolor[HTML]{D5E8D4}\textbf{2.70}
& \cellcolor[HTML]{D5E8D4}\textbf{75}
& \cellcolor[HTML]{D5E8D4}\textbf{106}
& \cellcolor[HTML]{D5E8D4}\textbf{122}
& \cellcolor[HTML]{D5E8D4}\textbf{207}
& \cellcolor[HTML]{D5E8D4}\textbf{145}
& \cellcolor[HTML]{D5E8D4}\textbf{131.0}
\\

\bottomrule
\end{tabular}
}

\vspace{-3mm}
\end{table*}

\paragraph{End-to-end efficiency.}
Table~\ref{tab:extended_main_results} further confirms that the throughput improvements of GrowPage translate into shorter workload completion time.
GrowPage achieves the lowest total inference time across all five benchmarks on both models.
Compared with Zipage, its average inference time decreases from 3.13 to 2.51 hours on DeepSeek-R1-Distill-Llama-8B and from 3.08 to 2.70 hours on Qwen3-8B.
The improvement becomes particularly pronounced on longer and more demanding workloads; for example, on LiveCodeBench with DeepSeek-R1-Distill-Llama-8B, GrowPage reduces the total inference time from 7.75 to 5.75 hours.
These results indicate that adapting per-request KV capacity reduces unnecessary memory residency and enables the serving engine to complete batched reasoning workloads more efficiently.

\paragraph{Per-token decoding efficiency.}
GrowPage also consistently achieves the lowest TPOT across all evaluated datasets.
Averaged over the five benchmarks, GrowPage reduces TPOT from 160.2 to 132.4 ms/token relative to Zipage on DeepSeek-R1-Distill-Llama-8B, and from 156.8 to 131.0 ms/token on Qwen3-8B.
This suggests that on-demand capacity control not only increases aggregate concurrency but also reduces the attention cost associated with unnecessarily large KV working sets.
Interestingly, MorphKV, R-KV, and G-KV achieve substantially lower TPOT than FullKV on several long-generation workloads, yet exhibit much lower aggregate TPS in Table~\ref{tab:main_results}.
This discrepancy highlights that reducing per-token computation alone does not necessarily translate into higher serving throughput: without system-level integration with continuous batching and paged memory management, algorithmic KV compression may fail to efficiently exploit the freed memory under high concurrency.
GrowPage combines both advantages, maintaining low per-token latency while translating KV savings into higher aggregate serving throughput.

\section{Additional Capacity-Control Analysis}
\label{app:capacity_control_analysis}

\subsection{Alternative Capacity-Control Policies}
\label{app:capacity_policy_ablation}

We further investigate whether the performance gains of GrowPage arise specifically from its demand-guided capacity control, rather than merely from changing the KV capacity or the overall frequency of compression and growth.
We compare the default GrowPage policy with four alternative capacity-control strategies on Qwen3-8B with AMC23.

\paragraph{Compared policies.}
\textsc{GrowPage-Fixed} disables capacity expansion and maintains a fixed KV budget of 4096 tokens; whenever the current capacity is exhausted, it applies \emph{Compress \& Hold}.
\textsc{GrowPage-Random} removes the dependence on $\Delta_t$ and randomly selects \emph{Compress \& Hold} with probability $0.7$ and \emph{Grow by One Page} otherwise.
The probability is chosen to approximately match the empirical compression frequency of the default GrowPage policy, providing a frequency-matched control for evaluating whether the timing of capacity decisions matters.
\textsc{GrowPage-Inverse} reverses the default interpretation of the demand signal: it grows when $\Delta_t \le 0$ and compresses when $\Delta_t>0$.
Finally, \textsc{GrowPage-Shrink} introduces explicit bidirectional resizing with three actions:
\begin{equation}
\pi_{\mathrm{shrink}}(\Delta_t)=
\begin{cases}
\text{\emph{Shrink}}, & \Delta_t < -0.005,\\
\text{\emph{Compress \& Hold}}, & -0.005 \le \Delta_t \le 0,\\
\text{\emph{Grow by One Page}}, & \Delta_t > 0.
\end{cases}
\label{eq:shrink_policy}
\end{equation}
Here, \emph{Shrink} compacts two pages worth of historical KV states and returns one physical page to the global memory pool, while leaving one page worth of free slots for subsequent decoding.
We impose a minimum KV capacity of 1024 tokens to prevent excessive contraction.
The threshold $-0.005$ is motivated by the low-demand region identified in the counterfactual analysis of Figure~\ref{fig:capacity_validation}.

\begin{table*}[t]
\centering
\caption{
Comparison of alternative capacity-control policies on Qwen3-8B with AMC23.
Total time denotes the end-to-end inference time of the workload.
Lower is better for Avg. KV Budget, TPOT, and Total Time.
}
\label{tab:capacity_policy_ablation}

\setlength{\tabcolsep}{7pt}
\small

\resizebox{1.0\linewidth}{!}{
\begin{tabular}{lccccc}
\toprule
Method
& Avg. KV Budget $\downarrow$
& Pass@1 (\%) $\uparrow$
& TPS (tokens/s) $\uparrow$
& TPOT (ms/token) $\downarrow$
& Total Time (h) $\downarrow$
\\
\midrule

\textsc{GrowPage-Fixed}
& 4096
& 89.9
& 1874
& 136
& 1.59
\\

\textsc{GrowPage-Shrink}
& \textbf{2099}
& 84.2
& \textbf{2546}
& \textbf{107}
& 1.44
\\

\textsc{GrowPage-Random}
& 3728
& 88.3
& 2008
& 137
& 1.57
\\

\textsc{GrowPage-Inverse}
& 6167
& 90.8
& 1340
& 175
& 2.13
\\

Zipage \texttt{{\scriptsize (ACL'26)}}
& 4096
& 89.5
& 1805
& 138
& 1.63
\\

\cellcolor[HTML]{D5E8D4}\textbf{GrowPage (Ours)}
& \cellcolor[HTML]{D5E8D4}3164
& \cellcolor[HTML]{D5E8D4}\textbf{91.4}
& \cellcolor[HTML]{D5E8D4}2261
& \cellcolor[HTML]{D5E8D4}122
& \cellcolor[HTML]{D5E8D4}\textbf{1.42}
\\

\bottomrule
\end{tabular}
}

\vspace{-2mm}
\end{table*}

\paragraph{Effect of demand-guided adaptation.}
Table~\ref{tab:capacity_policy_ablation} shows that dynamically adapting KV capacity is important for obtaining a favorable accuracy--efficiency trade-off.
Compared with \textsc{GrowPage-Fixed}, the default policy reduces the average KV budget from 4096 to 3164 tokens while improving TPS from 1874 to 2261 and Pass@1 from 89.9\% to 91.4\%.
This indicates that the benefit cannot be reproduced by simply maintaining a static capacity and repeatedly compressing within it.

More importantly, \textsc{GrowPage-Random} produces an action frequency close to that of the default policy, but performs substantially worse.
Despite using a similar fraction of compression and growth decisions, the random policy reaches only 88.3\% Pass@1 at 2008 tokens/s, compared with 91.4\% at 2261 tokens/s for GrowPage.
It also maintains a larger average KV budget of 3728 tokens.
This result shows that the gain does not arise merely from the marginal frequency of compression and growth; \emph{when} each action is applied is critical.

Reversing the demand signal leads to the opposite behavior.
\textsc{GrowPage-Inverse} expands much more aggressively, increasing the average KV budget to 6167 tokens and reducing throughput to 1340 tokens/s, while still achieving lower accuracy than the default GrowPage policy.
This result complements the counterfactual analysis in Figure~\ref{fig:capacity_validation} and confirms that the direction of $\Delta_t$ carries meaningful information for capacity allocation rather than serving only as a generic activity signal.

\paragraph{Why not aggressively shrink capacity?}
\textsc{GrowPage-Shrink} achieves the smallest average KV budget and the highest raw throughput, but its Pass@1 drops sharply to 84.2\%, $7.2$ percentage points below the default GrowPage policy.
The result exposes the cost of aggressively reclaiming physical pages from active reasoning requests: once additional historical KV states are irreversibly removed, later reasoning stages may again require information that is no longer available.
GrowPage therefore adopts the more conservative \emph{Compress \& Hold} operation in the main design, retaining the current physical allocation while creating room for subsequent tokens.
This yields a substantially better balance between memory efficiency, serving throughput, and reasoning accuracy.

\subsection{System-Level Event Analysis}
\label{app:capacity_event_analysis}

To better understand how different capacity-control policies affect the serving runtime, we further record the number of page-growth, compression, fallback, and request-preemption events.
A \emph{fallback} occurs when a requested page expansion cannot be safely satisfied by the global free-page pool and is therefore converted to compression.
A \emph{preemption} is triggered under severe memory pressure when page expansion fails and the subsequent fallback compression still cannot provide sufficient GPU memory; the scheduler then removes an active decoding request from the running batch, releases its occupied resources, and returns it to the waiting queue for later rescheduling.
For \textsc{GrowPage-Shrink}, we separately report explicit \emph{Shrink} events; these events are included in the total compression count recorded by the runtime but are separated from ordinary \emph{Compress \& Hold} events in Table~\ref{tab:capacity_event_analysis}.

\begin{table*}[t]
\centering
\caption{
Runtime capacity-control events on Qwen3-8B with AMC23.
Grow Ratio is computed over the primary capacity decisions before fallback handling.
For \textsc{GrowPage-Shrink}, explicit Shrink events are reported separately from Compress \& Hold.
}
\label{tab:capacity_event_analysis}

\setlength{\tabcolsep}{8pt}
\small

\resizebox{1.0\linewidth}{!}{
\begin{tabular}{lrrrrrr}
\toprule

Method
& Grow
& Grow Ratio (\%)
& Compress \& Hold
& Shrink
& Fallback
& Preemptions
\\

\midrule

\textsc{GrowPage-Fixed}
& 0
& 0.0
& 22285
& --
& 0
& 19
\\

\textsc{GrowPage-Shrink}
& 17909
& 45.5
& 12540
& 8927
& 103
& 4668
\\

\textsc{GrowPage-Random}
& 9345
& 30.4
& 21399
& --
& 944
& 1894
\\

\textsc{GrowPage-Inverse}
& 19881
& 66.7
& 9931
& --
& 286
& 2997
\\

\cellcolor[HTML]{D5E8D4}\textbf{GrowPage (Ours)}
& \cellcolor[HTML]{D5E8D4}9500
& \cellcolor[HTML]{D5E8D4}30.9
& \cellcolor[HTML]{D5E8D4}21203
& \cellcolor[HTML]{D5E8D4}--
& \cellcolor[HTML]{D5E8D4}597
& \cellcolor[HTML]{D5E8D4}1361
\\

\bottomrule
\end{tabular}
}

\vspace{-2mm}
\end{table*}

\paragraph{Decision frequency versus decision timing.}
The runtime statistics provide a particularly controlled comparison between GrowPage and \textsc{GrowPage-Random}.
The two policies have almost identical primary action distributions: GrowPage grows at 30.9\% of capacity boundaries, while the random policy grows at 30.4\%.
Their total numbers of primary capacity decisions are also nearly identical.
Nevertheless, GrowPage achieves substantially higher accuracy and throughput in Table~\ref{tab:capacity_policy_ablation}.
This frequency-matched comparison further demonstrates that the effectiveness of GrowPage comes from using $\Delta_t$ to determine \emph{where} expansion is needed, rather than merely choosing an appropriate global compression-to-growth ratio.
The demand-guided policy is also accompanied by fewer fallback and preemption events than its random counterpart.

\paragraph{Cost of dynamic capacity adaptation.}
Dynamic capacity adaptation also introduces additional interactions with the global memory manager.
Unlike \textsc{GrowPage-Fixed}, which never requests new physical pages and therefore incurs almost no fallback or preemption events, GrowPage triggers 597 fallbacks and 1361 preemptions as requests dynamically compete for additional KV capacity.
This represents a system-level cost of on-demand expansion under memory pressure.
Nevertheless, GrowPage still achieves higher throughput (2261 vs.\ 1874 tokens/s) and accuracy (91.4\% vs.\ 89.9\%) while using a smaller average KV budget (3164 vs.\ 4096), indicating that the benefit of demand-aware capacity allocation outweighs the additional scheduling pressure in our setting.

\paragraph{Capacity over-expansion and bidirectional resizing.}
The event distribution of \textsc{GrowPage-Inverse} further illustrates the importance of the direction of the demand signal.
Reversing the policy raises the growth ratio from 30.9\% to 66.7\%, leading to substantially larger average KV residency and lower serving throughput.
Although this policy retains more KV information, the additional capacity does not translate into higher reasoning accuracy, indicating that indiscriminate expansion wastes shared GPU memory.

\textsc{GrowPage-Shrink} exhibits a different failure mode.
It triggers 8927 explicit shrink operations while also performing 17909 subsequent growth operations, resulting in substantially more bidirectional capacity transitions.
This aggressive resizing is accompanied by 4668 request preemptions, considerably more than the 1361 observed with the default GrowPage policy.
Together with the significant accuracy degradation in Table~\ref{tab:capacity_policy_ablation}, these results suggest that aggressively reclaiming pages from live reasoning requests can introduce unnecessary capacity churn and irreversible information loss.
The default GrowPage policy therefore favors conservative \emph{Compress \& Hold} and incremental growth, which provides more stable online capacity adaptation under PagedAttention-based serving.

\section{Request-Level Capacity Alignment}
\label{app:request_capacity_alignment}

\begin{figure*}[t]
  \centering
  \includegraphics[width=0.98\linewidth, keepaspectratio]{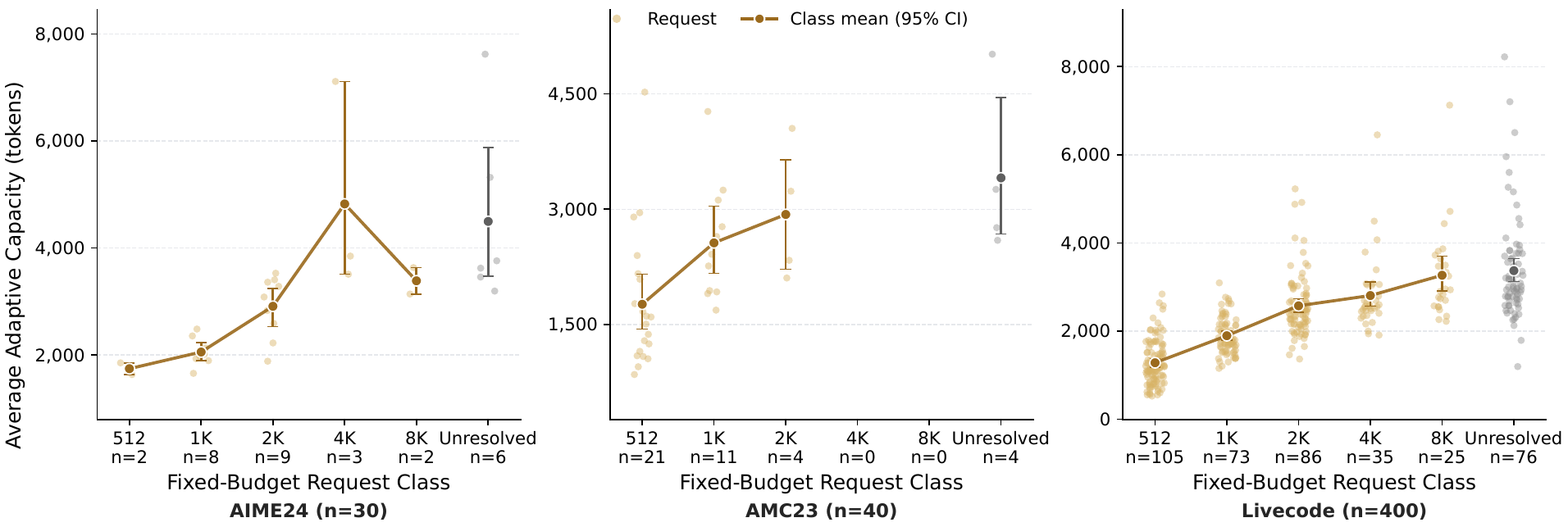}
\caption{
Alignment between minimum sufficient fixed-budget classes and GrowPage's adaptive KV capacity.
Requests are grouped by the minimum sufficient budget identified in Figure~\ref{fig:minimum_budgets}, and each point shows the average capacity allocated by GrowPage to the same request under the same random seed.
Markers and error bars denote the class mean and 95\% confidence interval.
}
\label{fig:adaptive_capacity_alignment}
\end{figure*}

Figure~\ref{fig:minimum_budgets} shows that the minimum sufficient fixed KV budget varies substantially across requests.
We further examine whether GrowPage's online capacity allocation reflects this request-level demand heterogeneity.
Specifically, we group requests according to the minimum sufficient budget $B_i^\star$ identified in Figure~\ref{fig:minimum_budgets}, and rerun the same requests with GrowPage using the same random seed.
For each request, we record its average allocated KV capacity over the decoding trajectory.
Figure~\ref{fig:adaptive_capacity_alignment} reports the resulting request-level capacities together with the mean and 95\% confidence interval for each fixed-budget class.

Across the populated budget classes, requests requiring larger fixed budgets generally receive larger adaptive capacities from GrowPage.
This trend is particularly clear on LiveCodeBench, where the average adaptive capacity increases consistently from the 512-token class to the 8K-token class.
AMC23 shows the same tendency over its populated 512, 1K, and 2K classes.
AIME24 also exhibits an overall positive relationship, although the sparsely populated 4K and 8K classes show larger variance.
These results indicate that GrowPage does not allocate capacity uniformly across requests, but instead adapts its runtime KV allocation in accordance with their underlying memory demands.

Importantly, the adaptive capacity is not expected to numerically match $B_i^\star$.
The latter denotes the minimum \emph{fixed} capacity required throughout an entire generation, whereas GrowPage changes the allocated capacity online as demand evolves.
Consequently, a request belonging to a high-budget class can still maintain a substantially smaller average capacity by using additional pages only during high-demand stages.
This distinction highlights the advantage of on-demand budgeting over static reservation.

We also observe that requests categorized as \emph{Unresolved} tend to receive relatively large adaptive capacities.
Since these requests are not consistently solved under any tested fixed budget, this behavior suggests that GrowPage responds to their elevated attention demand rather than to the eventual correctness outcome.
Overall, the alignment between $B_i^\star$ and GrowPage's adaptive allocation provides request-level evidence that the proposed demand signal translates heterogeneous reasoning demands into differentiated KV capacity.

\end{document}